\documentclass[10pt,journal,compsoc]{IEEEtran}

\ifCLASSOPTIONcompsoc
  \usepackage[nocompress]{cite}
\else
  \usepackage{cite}
\fi

\ifCLASSINFOpdf
\else
  
\fi

\usepackage[dvips]{graphicx}
\usepackage{amsfonts}
\usepackage{amsmath}
\usepackage{amssymb}
\usepackage{amsthm}
\usepackage{dsfont}
\usepackage{mathabx}
\usepackage{bm}
\usepackage{pstricks}
\usepackage{algpseudocode}

\ifCLASSOPTIONcompsoc
  \usepackage[caption=false,font=footnotesize,labelfont=sf,textfont=sf]{subfig}
\else
  \usepackage[caption=false,font=footnotesize]{subfig}
\fi

\newtheorem{definition}{Definition}
\newtheorem{proposition}{Proposition}
\newtheorem{lemma}{Lemma}
\newtheorem{theorem}{Theorem}

\newcommand{\numVar}{\ensuremath{N}}

\newcommand{\e}{\ensuremath{\mathbf{e}}}
\newcommand{\X}{\ensuremath{\mathbf{X}}}
\newcommand{\x}{\ensuremath{\mathbf{x}}}
\newcommand{\Y}{\ensuremath{\mathbf{Y}}}
\newcommand{\y}{\ensuremath{\mathbf{y}}}
\newcommand{\Z}{\ensuremath{\mathbf{Z}}}
\newcommand{\z}{\ensuremath{\mathbf{z}}}

\newcommand{\xs}{\mathcal{X}}
\newcommand{\XS}{\bm{\xs}}
\newcommand{\ys}{\mathcal{Y}}
\newcommand{\YS}{\bm{\ys}}
\newcommand{\zs}{\mathcal{Z}}
\newcommand{\ZS}{\bm{\zs}}

\newcommand{\SA}{\ensuremath{\mathcal{A}}}
\newcommand{\HC}{\ensuremath{\mathcal{H}}}

\newcommand{\val}{\ensuremath{\mathbf{val}}}
\newcommand{\pa}{\ensuremath{\mathbf{pa}}}
\newcommand{\ch}{\ensuremath{\mathbf{ch}}}
\newcommand{\desc}{\ensuremath{\mathbf{desc}}}
\newcommand{\anc}{\ensuremath{\mathbf{anc}}}
\newcommand{\scope}{\ensuremath{\mathbf{sc}}}
\newcommand{\Node}{\mathsf{N}}
\newcommand{\SumNode}{\mathsf{S}}
\newcommand{\ProdNode}{\mathsf{P}}
\newcommand{\DistNode}{\mathsf{D}}
\newcommand{\DistNodes}{\bm{\mathsf{D}}}
\newcommand{\SumNodes}{\bm{\mathsf{S}}}
\newcommand{\ProdNodes}{\bm{\mathsf{P}}}
\newcommand{\Child}{\mathsf{C}}
\newcommand{\Parent}{\mathsf{F}}
\newcommand{\w}{w}
\newcommand{\graph}{\ensuremath{\mathcal{G}}}
\newcommand{\SPN}{\ensuremath{\mathcal{S}}}

\newcommand{\IV}[2]{\lambda_{#1=#2}}

\newcommand{\p}{p}
\newcommand{\cbar}{\,|\,}
\newcommand{\sbar}{\ensuremath{\,|\,}}
\newcommand{\subv}[2]{#1{[#2]}}

\newcommand{\Sanc}{\ensuremath{{\mathbf{anc}_{\bm\SumNode}}}}
\newcommand{\Sdesc}{\ensuremath{{\mathbf{desc}_{\bm\SumNode}}}}

\newcommand{\aug}{\ensuremath{\mathbf{aug}}}

\newcommand{\numSamples}{\ensuremath{L}}
\newcommand{\sampC}{\ensuremath{l}}

\begin{document}
\title{On the Latent Variable Interpretation in Sum-Product Networks}

\author{Robert Peharz, Robert Gens, Franz Pernkopf,~\IEEEmembership{Senior Member,~IEEE,} and Pedro Domingos
\IEEEcompsocitemizethanks{\IEEEcompsocthanksitem R.~Peharz is with the Institute of Physiology (iDN), Medical University of Graz and BioTechMed--Graz.\protect\\
E-mail: robert.peharz@gmail.com
\IEEEcompsocthanksitem R.~Gens and P.~Domingos are with the Department of Computer Science and Engineering, 
University of Washington.\protect\\
E-mail: rcg@cs.washington.edu, pedrod@cs.washington.edu
\IEEEcompsocthanksitem F.~Pernkopf is with the Signal Processing and Speech Communication Lab, 
Graz University of Technology.\protect\\
E-mail: pernkopf@tugraz.at}
\thanks{Manuscript received November 12, 2015; revised June 20, 2016; accepted September 30, 2016.}}

\markboth{IEEE Transactions on Pattern Analysis and Machine Intelligence, accepted pre-print version, October 2016}%
{Peharz \MakeLowercase{\textit{et al.}}: On the Latent Variable Interpretation in Sum-Product Networks}

\IEEEtitleabstractindextext{%
\begin{abstract}
One of the central themes in Sum-Product networks (SPNs) is the interpretation of sum nodes as marginalized latent
variables (LVs). This interpretation yields an increased syntactic or semantic structure, allows the application of the EM algorithm and
to efficiently perform MPE inference. In literature, the LV interpretation was justified by explicitly introducing the indicator variables
corresponding to the LVs' states. However, as pointed out in this paper, this approach is in conflict with the completeness condition in
SPNs and does not fully specify the probabilistic model. We propose a remedy for this problem by modifying the original approach for
introducing the LVs, which we call SPN augmentation. We discuss conditional independencies in augmented SPNs, formally establish
the probabilistic interpretation of the sum-weights and give an interpretation of augmented SPNs as Bayesian networks. Based on
these results, we find a sound derivation of the EM algorithm for SPNs. 
Furthermore, the Viterbi-style algorithm for MPE proposed in literature was never proven to be correct. We show that this is indeed a correct algorithm,
when applied to selective SPNs, and in particular when applied to augmented SPNs.
Our theoretical results are confirmed in experiments on synthetic data and 103 real-world datasets.
\end{abstract}

}

\maketitle

\IEEEdisplaynontitleabstractindextext

\IEEEpeerreviewmaketitle

\ifCLASSOPTIONcompsoc
\IEEEraisesectionheading{\section{Introduction}\label{sec:introduction}}
\else
\section{Introduction}
\label{sec:introduction}
\fi

\IEEEPARstart{S}{um-Product Networks} are a promising type of probabilistic model, combining the domains of
deep learning and graphical models \cite{Poon2011, Gens2013}. 
One of their main advantages is that many interesting inference scenarios are expressed as single forward 
and/or backward passes, i.e. these inference scenarios have a computational cost linear in the SPN's 
representation size. 
SPNs have shown convincing performance in applications such as image completion \cite{Poon2011, Dennis2012, Peharz2013}, 
computer vision \cite{Amer2012}, classification \cite{Gens2012} and speech and language modeling \cite{Peharz2014, Cheng2014, Zhoerer2015}. 
Since their proposition \cite{Poon2011}, one of the central themes in SPNs has been their interpretation as 
hierarchically structured latent variable (LV) models. 
This is essentially the same approach as the LV interpretation in mixture models. 
Consider for example a Gaussian mixture model with K components over a set of random variables (RVs) $\X$:
\begin{equation}
\p(\X) = \sum_{k=1}^K \w_k \, \mathcal{N}(\X \cbar \bm{\mu}_k, \bm{\Sigma}_k), 
\end{equation}
where $\mathcal{N}(\cdot \cbar \cdot)$ is the Gaussian PDF, $\bm{\mu}_k$ and $\bm{\Sigma}_k$ are the means and covariances of the $k^\text{th}$ component, and $\w_k$ are the mixture weights with $\w_k \geq 0$, $\sum \w_k = 1$.
The GMM can be interpreted in two ways:
i) It is a convex combination of PDFs and thus itself a PDF, or
ii) it is a marginal distribution of a distribution $\p(\X,Z)$ over $\X$ and a latent, marginalized variable $Z$, where $\p(\X \cbar Z = k) = \mathcal{N}(\X \cbar \bm{\mu}_k, \bm{\Sigma}_k)$ and $\p(Z = k) = \w_k$.
The second interpretation, the LV interpretation, yields a syntactically well-structured model.
For example, following the LV interpretation, it is clear how to draw samples from $\p(\X)$ by using ancestral sampling.
This structure can also be of semantic nature, for instance when $Z$ represents a clustering of $\X$ or when $Z$ is a class variable.
Furthermore, the LV interpretation allows the application of the EM algorithm -- which is essentially maximum-likelihood learning under missing data
\cite{Dempster1977, Ghahramani1994} -- and enables advanced Bayesian techniques \cite{Lee2014, Trapp2016}.

Mixture models can be seen as a special case of SPNs with a single sum node, which corresponds to a single LV.
More generally, SPNs can have arbitrarily many sum nodes, each corresponding to its own LV, leading to a hierarchically structured model.
In \cite{Poon2011}, the LV interpretation in SPNs was justified by explicitly introducing the LVs in the SPN model, using the so-called \emph{indicator variables} corresponding to the LVs' states.
However, as shown in this paper, this justification is actually too simplistic, since it is potentially in conflict with the \emph{completeness} condition \cite{Poon2011}, leading to an incompletely specified model.
As a remedy we propose the \emph{augmentation} of an SPN, which additionally to the IVs also introduces the so-called \emph{twin sum nodes}, in order to completely specify the LV model.
We further investigate the independency structure of the LV model resulting from augmentation and find a parallel to the local independence assertions in Bayesian networks (BNs) \cite{Pearl1988, Koller2009}.
This allows us to define a BN representation of the augmented SPN.
Using our BN interpretation and the differential approach \cite{Darwiche2003, Peharz2015} in augmented SPNs, we give a sound derivation of the (soft) EM algorithm for SPNs.

Closely related to the LV interpretation is the inference scenario of finding the most-probable-explanation (MPE), i.e.~finding a probability maximizing assignment for all RVs.
Using results form \cite{deCampos2011, Peharz2015b}, we first point out that this problem is generally NP-hard for SPNs.
In \cite{Poon2011} it was proposed that an MPE solution can be found efficiently
when maximizing over both model RVs (i.e. non-latent RVs) and LVs. 
The proposed algorithm replaces sum nodes by max nodes and recovers the solution by using Viterbi-style backtracking. 
However, it was not shown that this algorithm delivers a correct MPE solution. 
In this paper, we show that this algorithm is indeed correct, \emph{when applied to selective SPNs} \cite{Peharz2014b}. 
In particular, since augmented SPNs are selective, this algorithm obtains an MPE solution in augmented SPNs.
However, when applied to \emph{non-augmented} SPNs, the algorithm still returns an MPE solution of the augmented SPN, but implicitly 
assumes that the weights for all twin sums are deterministic, i.e.~they are all 0 except a single 1. 
This leads to a phenomenon in MPE inference which we call \emph{low-depth bias}, i.e. more shallow parts of the SPN are 
preferred during backtracking.

The main contribution in this paper is to provide a sound theoretical foundation for the LV interpretation in SPNs and
related concepts, i.e. the EM algorithm and MPE inference.
Our theoretical findings are confirmed in experiments on synthetic data and 103 real-world datasets.

The paper is organized as follows: In the remainder of this section we introduce notation, review SPNs and discuss 
related work. 
In Section 2 we propose the augmentation of SPNs, show its soundness as hierarchical LV model and give an interpretation 
as BN. 
Furthermore, we discuss independency properties in augmented SPNs and the interpretation of sum-weights as conditional probabilities. 
The EM algorithm for SPNs is derived in Section 3. In Section 4 we discuss MPE inference for SPNs. 
Experiments are presented in Section 5 and Section 6 concludes the paper. 
Proofs for our theoretical findings are deferred to the Appendix.

\subsection{Background and Notation}   \label{sec:backgroundNotation}
RVs are denoted by upper-case letters $W$, $X$, $Y$ and $Z$.
The set of values of an RV $X$ is denoted by $\val(X)$, where corresponding lower-case letters denote elements of $\val(X)$, e.g.~$x$ is an element of $\val(X)$.
Sets of RVs are denoted by boldface letters $\mathbf{W}$, $\X$, $\Y$ and $\Z$.
For RV set $\X = \{X_1,\dots,X_N\}$, we define $\val(\X) = \bigtimes_{n=1}^N \val(X_n)$ and use corresponding lower-case boldface letters for elements of $\val(\X)$, e.g.~$\x$ is an element of $\val(\X)$.
For a sub-set $\Y \subseteq \X$, $\subv{\x}{\Y}$ denotes the projection of $\x$ onto $\Y$.

The elements of $\val(\X)$ can be interpreted as \emph{complete} evidence, assigning each RV in $\X$ a fixed value.
\emph{Partial} evidence about $X$ is represented as a subset $\xs \subseteq \val(X)$, which is an element of the sigma-algebra $\SA_X$ induced by RV $X$.
For all RVs we use $\SA_X = \{ \xs \in \bm{\mathcal{B}} ~|~ \xs \subseteq \val(X) \}$, $\bm{\mathcal{B}}$ being the Borel-sets over $\mathbb{R}$.
For discrete RVs, this choice yields the power-set $\SA_X = 2^{\val(X)}$.
For example, partial evidence $\xs = \{1,3,5\}$ for a discrete RV $X$ with $\val(X) = \{1,\dots,6\}$ represents evidence
that $X$ takes one of the states $1$, $3$ or $5$, and $\ys = [-\infty,\pi]$ for a real-valued RV $Y$ represents evidence that $Y$ takes a value
smaller than $\pi$. 
Formally speaking, partial evidence is used to express the domain of \emph{marginalization or maximization} for a particular RV.

For sets of RVs $\X = \{X_1, \dots, X_\numVar\}$, we use the product sets
$\HC_\X := \{ \bigtimes_{n=1}^\numVar \xs_{n} ~|~ \xs_{n} \in \SA_{X_n} \}$ to represent partial evidence about $\X$.
Elements of $\HC_\X$ are denoted using boldface notation, e.g.~$\XS$.
When $\Y \subseteq \X$ and $\XS \in \mathcal{H}_\X$, we define $\subv{\bm{\mathcal{X}}}{\Y} := \{\subv{\x}{\Y} ~|~ \x \in \bm{\mathcal{X}}\}$.
Furthermore, we use $\e$ to symbolize any combination of complete and partial evidence, i.e.~for RVs $\X$ we have some complete evidence $\x'$ for $\X' \subseteq \X$ and some partial evidence $\XS'' \in \HC_{\X''}$ for $\X'' = \X \setminus \X'$.

Given a node $\Node$ in some directed graph $\graph$, let $\ch(\Node)$ and $\pa(\Node)$ be the set of children and parents of $\Node$, respectively.
Furthermore, let $\desc(\Node)$ be the set of descendants of $\Node$, recursively defined as the set containing $\Node$ itself and any child of a descendant.
Similarly, we define $\anc(\Node)$ as the ancestors of $\Node$, recursively defined as the set containing $\Node$ itself and any parent of an ancestor.
SPNs are defined as follows.
\begin{definition}[Sum-Product Network]
\label{def:SPN}
A Sum-Product network (SPN) $\SPN$ over a set of RVs $\X$ is a tuple $(\graph, \bm{\w})$ where $\graph$ is a connected, rooted and acyclic directed graph, and 
$\bm{\w}$ is a set of non-negative parameters.
The graph $\graph$ contains three types of nodes: distributions, sums and products.
All \emph{leaves} of $\graph$ are distributions and all \emph{internal} nodes are either sums or products.
A distribution node (also called input distribution or simply distribution) $\DistNode_\Y\colon \val(\Y) \mapsto [0,\infty]$ is a distribution function over a subset of RVs $\Y \subseteq \X$, i.e.~either a PMF (discrete RVs), a PDF (continuous RVs), or a mixed distribution function (discrete and continuous RVs mixed).
A sum node $\SumNode$ computes a weighted sum of its children, i.e.
$\SumNode = \sum_{\Child \in \ch(\SumNode)} \w_{\SumNode, \Child} \, \Child$, 
where $\w_{\SumNode, \Child}$ is a non-negative weight associated with edge $\SumNode \rightarrow \Child$, and $\bm{\w}$ contains the weights for all outgoing sum-edges.
A product node $\ProdNode$ computes the product over its children, i.e.
$\ProdNode = \prod_{\Child \in \ch(\ProdNode)} \Child$.
The sets $\SumNodes(\SPN)$ and $\ProdNodes(\SPN)$ contain all sum nodes and all product nodes in $\SPN$, respectively.

The \emph{size} $|\SPN|$ of the SPN is defined as the number of nodes and edges in $\graph$.
For any node $\Node$ in $\graph$, the \emph{scope} of $\Node$ is defined as
\begin{equation}
\scope(\Node) = 
\begin{cases}
\Y & \text{if } \Node \text{ is a distribution } \DistNode_\Y \\
\bigcup_{\Child \in \ch(\Node)} \scope(\Child) & \text{otherwise.}
\end{cases}
\end{equation}
The function computed by $\SPN$ is the function computed by its root and denoted as $\SPN(\x)$, where without loss of generality we assume that the scope of the root is $\X$.
\end{definition}
We use symbols $\DistNode$, $\SumNode$, $\ProdNode$, $\Node$, $\Child$ and $\Parent$ for nodes in SPNs, where $\DistNode$ denotes a distribution, $\SumNode$ denotes a sum, and $\ProdNode$ denotes a product.
Symbols $\Node$, $\Child$ and $\Parent$ denote generic nodes, where $\Child$ and $\Parent$ indicate a child or parent relationship to another node, respectively.
The \emph{distribution} $\p_\SPN$ of an SPN $\SPN$ is defined as the normalized output of $\SPN$, i.e.~$\p_\SPN(\x) \propto \SPN(\x) $. 
For each node $\Node$, we define the \emph{sub-SPN} $\SPN_\Node$ rooted at $\Node$ as the SPN defined by the graph induced by the descendants of $\Node$ and the corresponding parameters.

Inference in unconstrained SPNs is generally intractable.
However, efficient inference in SPNs is enabled by two structural constraints, \emph{completeness} and \emph{decomposability} \cite{Poon2011}.
An SPN is 
\emph{complete} if for all sums $\SumNode$ it holds that 
\begin{equation}
\forall \Child',\Child'' \in \ch(\SumNode) \colon \scope(\Child') = \scope(\Child'').
\end{equation}
An SPN is \emph{decomposable} if for all products $\ProdNode$ it holds that 
\begin{equation}
\forall \Child',\Child'' \in \ch(\ProdNode), \Child' \not= \Child''\colon \scope(\Child') \cap \scope(\Child'') = \emptyset.
\end{equation}
Furthermore, a sum node $\SumNode$ is called \emph{selective} \cite{Peharz2014b} if for all choices of sum-weights $\bm{\w}$ and all possible inputs $\x$
it holds that at most one child of $\SumNode$ is non-zero.
An SPN $\SPN$ is called selective if all its sum nodes are selective.

As shown in \cite{Peharz2015,Peharz2015b}, integrating $\SPN(\x)$ over arbitrary sets $\XS \in \HC_\X$, i.e.~\emph{marginalization} over $\XS$, reduces to the corresponding integrals at the input distributions and evaluating sums and products in the usual way.
This property is known as \emph{validity} of the SPNs \cite{Poon2011}, and key for efficient inference.
In this paper we only consider complete and decomposable SPNs.
Without loss of generality \cite{Peharz2015, Zhao2015}, we assume \emph{locally normalized} sum-weights, i.e.~for each sum node $\SumNode$ we have $\sum_{\Child \in \ch(\SumNode)} \w_{\SumNode, \Child} = 1$, and thus $\p_\SPN \equiv \SPN$, i.e.~the SPN's normalization constant is $1$.

For RVs with finitely many states, we will use so-called \emph{indicator variables} (IVs) as input distributions \cite{Poon2011}.
For a finite-state RV $X$ and state $x \in \val(X)$, we introduce the IV $\IV{X}{x}(x') := \mathds{1}(x = x')$, assigning all probability mass to $x$.
A complete and decomposable SPN represents the \emph{(extended) network polynomial} of $\p_\SPN$, which can be used in the \emph{differential approach to inference} \cite{Darwiche2003,Poon2011,Peharz2015}.
Assume any evidence $\e$ which is evaluated in the SPN.
The derivatives of the SPN function with respect to the IVs (by interpreting the IVs as real-valued variables, see \cite{Darwiche2003,Peharz2015} for details) yield
\begin{equation}
\label{eq:diffapproach}
\frac{\partial {\SPN(\mathbf{e})}}{\partial \IV{X}{x}} = \SPN(X=x, \mathbf{e} \setminus X),
\end{equation}
representing the inference scenario of \emph{modified evidence}, i.e.~evidence $\mathbf{e}$ is modified such that $X$ is set to $x$.
The computationally attractive feature of the differential approach is that \eqref{eq:diffapproach} can be evaluated for \emph{all} $X\in\X$ and \emph{all} $x\in \val(X)$ simultaneously using a \emph{single} back-propagation pass in the SPN, after evidence has been evaluated.
Similarly, for the second (and higher) derivatives, we get
\begin{equation}
\label{eq:diffapproach2}
\frac{\partial^2 {\SPN(\mathbf{e})}}{\partial \IV{X}{x} \IV{Y}{y}} = 
\begin{cases}
\SPN(X=x, Y=y, \mathbf{e} \setminus \{X,Y\}) & \text{if } X \not= Y \\
0 & \text{otherwise}.
\end{cases}
\end{equation}
Furthermore, the differential approach can be generalized to SPNs with arbitrary input distributions, 
i.e. SPNs over RVs with countably infinite or uncountably many states (cf.~\cite{Peharz2015} for details).

\subsection{Related Work}
SPNs are related to negation normal forms (NNFs), a potential deep network representation of propositional
theories \cite{Darwiche1999, Darwiche2001, Darwiche2002}. 
Like in SPNs, structural constraints in NNFs enable certain polynomial-time queries in the represented theory. 
In particular, the notions of smoothness, decomposability and determinism in NNFs translate to the notions of completeness, decomposability and selectivity in SPNs, respectively.
The work on NNFs led to the concept of network polynomials as a multilinear representation of BNs over finitely many states 
\cite{Darwiche2003}, \cite{Darwiche2002b}. 
BNs were cast into an intermediate d-DNNF (deterministic decomposable NNF) representation in order to generate an arithmetic circuit (ACs), representing the
BN’s network polynomial. ACs, when restricted to sums and products, are equivalent to SPNs but have a slightly different syntax. 
In \cite{Lowd2008}, ACs were learned by optimizing an objective trading off the log-likelihood on the training set and the inference cost of the AC, measured as the worst-case number of arithmetic operations required for inference (i.e. the number of edges in the AC). 
The learned models still represent BNs with context-specific independencies \cite{Boutilier1996}. 
A similar approach learning Markov networks represented by ACs is followed in \cite{Lowd2013}. 
SPNs were the first time proposed in \cite{Poon2011}, where the represented distribution was not defined via a background graphical model any more, 
but directly as the normalized output of the network. 
In this work, SPNs were applied to image data, where a generic architecture reminiscent to convolutional neural networks was proposed. 
Structure learning algorithms not restricted to the image domain were proposed in \cite{Dennis2012, Gens2013, Peharz2013, Rooshenas2014, Adel2015, Vergari2015}.
Discriminative learning of SPNs, optimizing conditional likelihood, was proposed in \cite{Gens2012}.
Furthermore, there is a growing body of literature on theoretical aspects of SPNs and their relationship to other types of probabilistic models. 
In \cite{Delalleau2011} two families of functions were identified which are efficiently representable by deep, but not by shallow SPNs, where an SPN is considered as shallow if it has no more than three layers. 
In \cite{Peharz2015} it was shown that SPNs can w.l.o.g.~be assumed to be locally normalized and that the notion of consistency does not allow exponentially more compact models than decomposability. 
These results were independently found in \cite{Zhao2015}. 
Furthermore, in \cite{Peharz2015}, a sound derivation of inference mechanisms for generalized SPNs was given, i.e. SPNs over RVs with (uncountably)
infinitely many states. 
In \cite{Zhao2015}, a BN representation of SPNs was found, where LVs associated with sum nodes and the model RVs are organized in a two layer bipartite structure.
The actual SPN structure is captured in structured conditional probability tables (CPTs) using algebraic decision diagrams.
Recently, the notion of SPNs was generalized to sum-product functions over arbitrary semirings \cite{Friesen2016}.
This yields a general unifying framework for learning and inference, subsuming, among others, SPNs for probabilistic modeling, NNFs for logical propositions and function representations for integration and optimization.

\section{Latent Variable Interpretation}    \label{sec:latentVar}
\begin{figure}
\centering
 \subfloat[]{\includegraphics[scale=0.165]{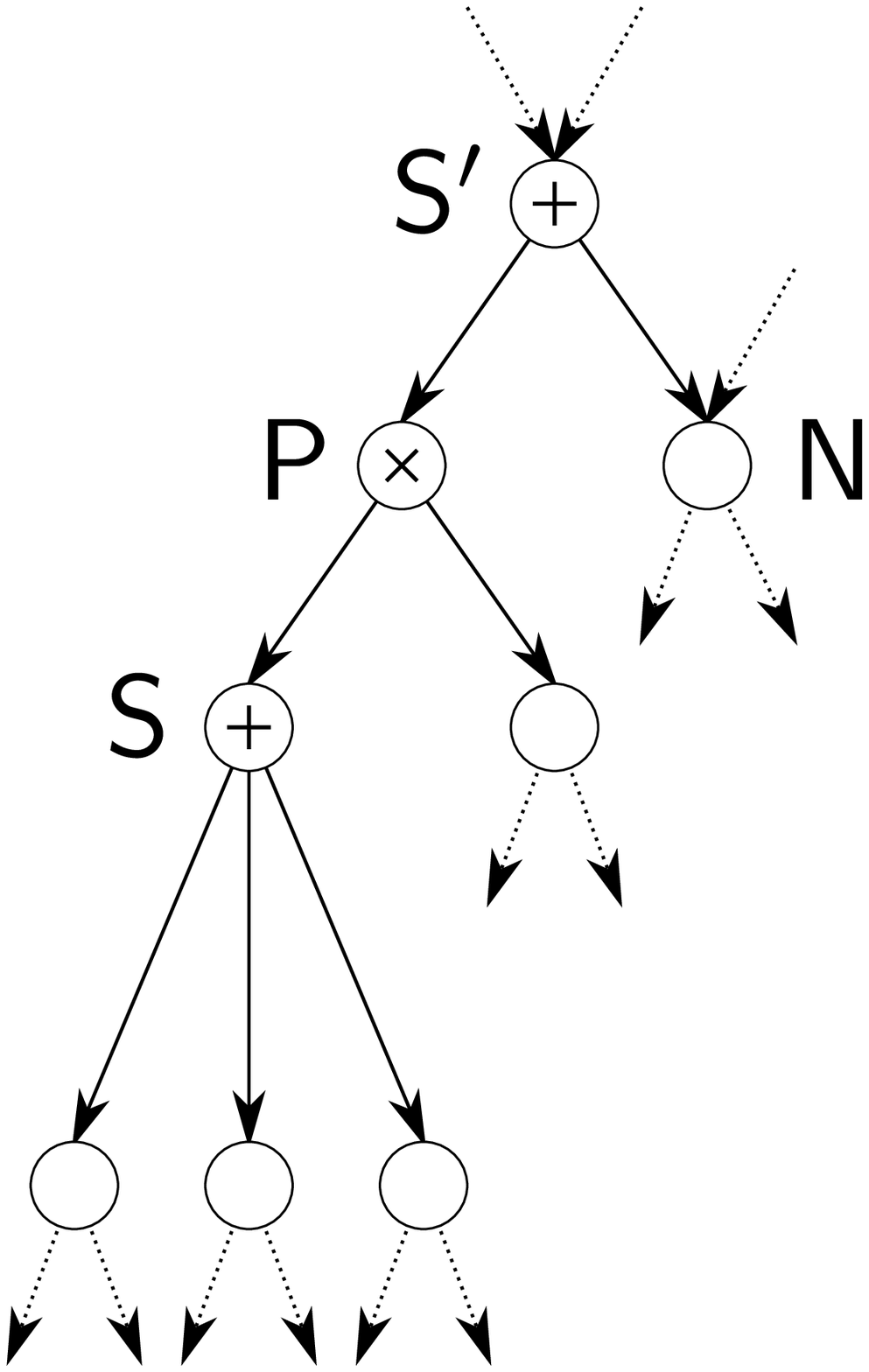}
\label{fig:problemAugmentSPNnaiv_1} 
  }
\subfloat[]{
  \includegraphics[scale=0.165]{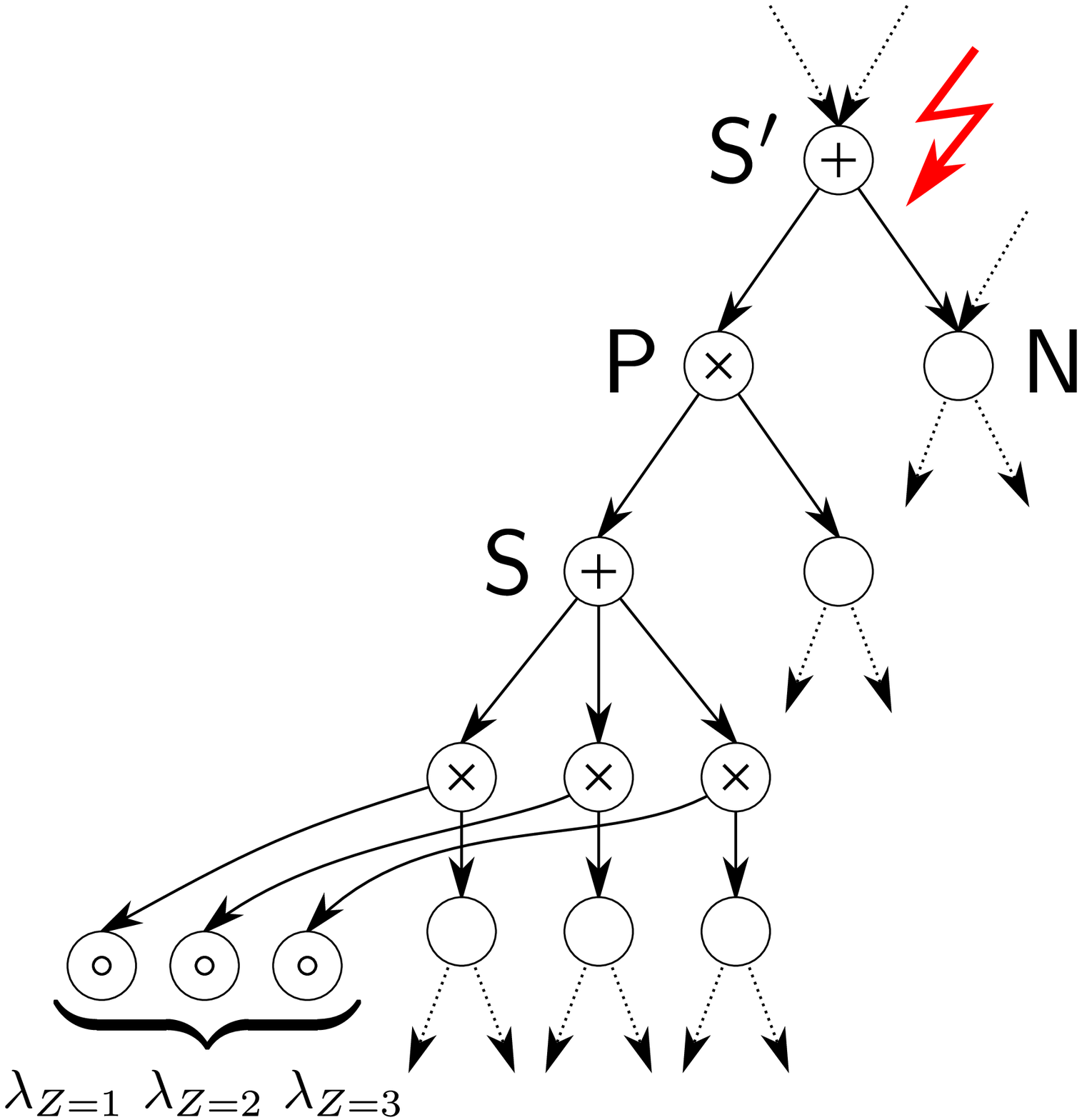}
	\label{fig:problemAugmentSPNnaiv_2}
 }
\subfloat[]{
  \includegraphics[scale=0.165]{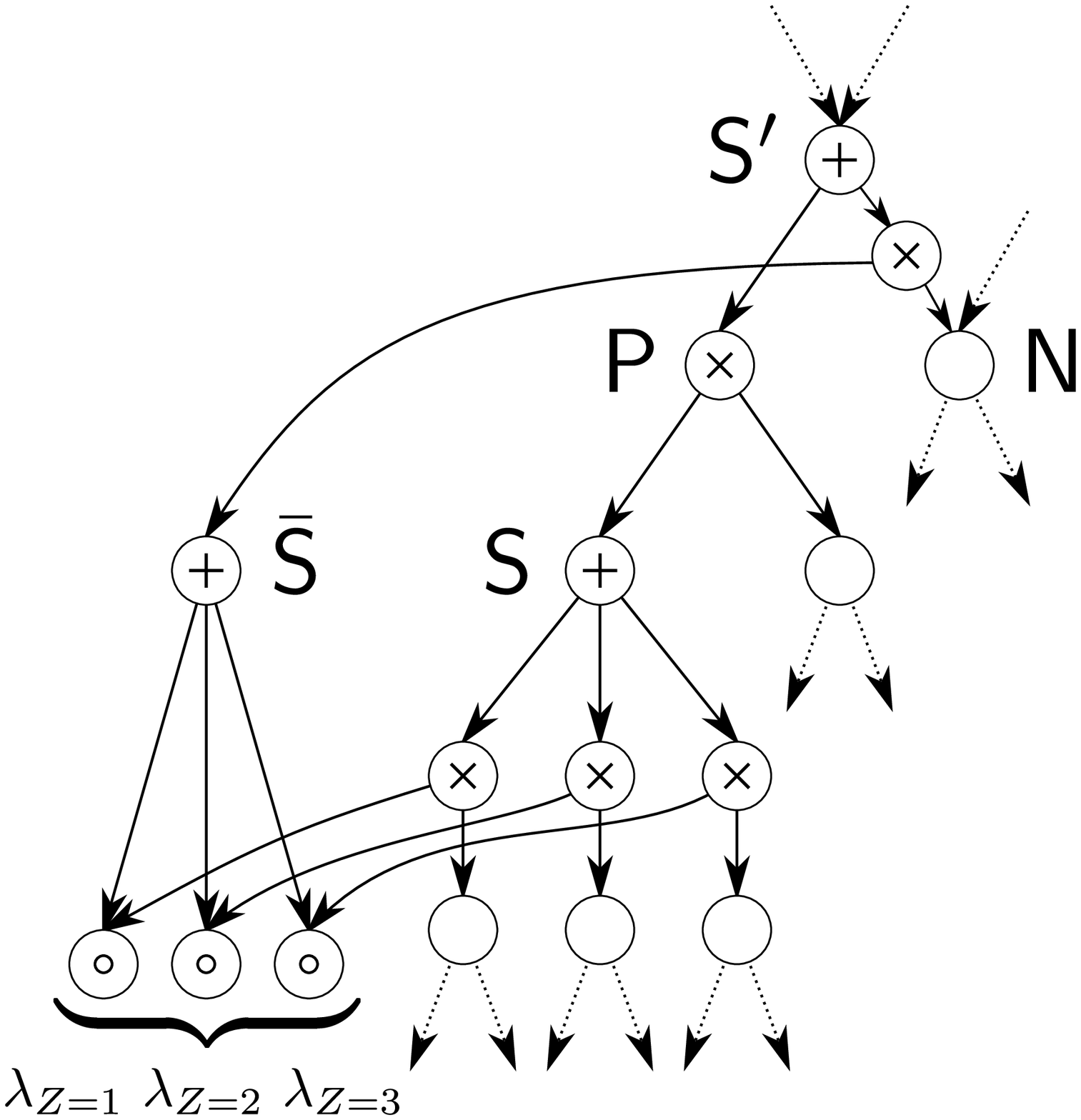}
	\label{fig:problemAugmentSPNnaiv_3} 
 }
 \caption{Problems occurring when IVs of LVs are introduced. 
\protect\subref{fig:problemAugmentSPNnaiv_1}: Excerpt of SPN containing a sum $\SumNode$, corresponding to LV $Z$. 
\protect\subref{fig:problemAugmentSPNnaiv_2}: Introducing IVs for $Z$ renders $\SumNode'$ incomplete, assuming that $\SumNode \notin \desc(\Node)$.
\protect\subref{fig:problemAugmentSPNnaiv_3}: Remedy by extending SPN further, introducing twin sum node $\bar{\SumNode}$.}
\label{fig:problemAugmentSPNnaiv}
\end{figure}

As pointed out in \cite{Poon2011}, each sum node in an SPN can be
interpreted as a marginalized LV, similar as in the GMM
example in Section 1. 
For each sum node $\SumNode$, one postulates a discrete LV $Z$ whose states correspond to the children of $\SumNode$. 
For each state, an IV and a product is introduced, such that the children are switched on/off by the corresponding IVs, 
as illustrated in Fig.~\ref{fig:problemAugmentSPNnaiv}.\footnote{In graphical representations of SPNs, IVs are depicted as nodes
containing a small circle, general distributions as nodes containing a Gaussian-like PDF, and sum and products as nodes with $+$ and $\times$ symbols. 
Empty nodes are of arbitrary type.} 
When all IVs in Fig.~\ref{fig:problemAugmentSPNnaiv_2} are set to 1, $\SumNode$ still computes the same value as in Fig.~\ref{fig:problemAugmentSPNnaiv_1}. 
Since setting all IVs of $Z$ to 1 corresponds to marginalizing $Z$, the sum $\SumNode$ should be interpreted as a latent, marginalized RV.

However, when we regard a larger structural context in Fig.~\ref{fig:problemAugmentSPNnaiv_2}, 
we recognize that this justification is actually too simplistic. 
Explicitly introducing the IVs renders the ancestor $\SumNode'$ incomplete, when $\SumNode$ is no descendant of $\Node$, and
$Z$ is thus not in the scope of $\Node$. 
Note that setting all IVs to 1 in an \emph{incomplete} SPN generally does \emph{not} correspond to marginalization. 
Furthermore, note that also $\SumNode'$ corresponds to an LV, say $Z'$. 
While we know the probability distribution of $Z$ if $Z'$ is in the state corresponding to $\ProdNode$, namely the weights of $\SumNode$, we do not know this distribution when $Z'$ is in the state corresponding to $\Node$. 
Intuitively, we recognize that the state of $Z$ is “irrelevant” in this case, since it does not influence the resulting distribution over the model RVs $\X$.
Nevertheless, the probabilistic model is not completely specified, which is unsatisfying.

A remedy for these problems is shown in Fig.~\ref{fig:problemAugmentSPNnaiv_3}.
We introduce the twin sum node $\bar \SumNode$ whose children are the IVs corresponding to $Z$. 
The twin $\bar \SumNode$ is connected as child of an additional product node, which is interconnected between $\SumNode'$ and $\Node$. 
Since this new product node has scope $\scope(\Node) \cup \{Z\}$, $\SumNode'$ is rendered complete now. 
Furthermore, if $Z'$ takes the state corresponding to $\Node$ (or actually the state corresponding to the new product node), we now have a specified conditional distribution for $Z$, namely the weights of the twin sum node. 
Clearly, given that all IVs of $Z$ are set to 1, the network depicted in Fig.~\ref{fig:problemAugmentSPNnaiv_3} still computes the same function as the network in Fig.~\ref{fig:problemAugmentSPNnaiv_1} (or Fig.~\ref{fig:problemAugmentSPNnaiv_2}), since $\bar \SumNode$ constantly outputs 1, as long as we use normalized weights for it. 
Which weights should be used for the twin sum node $\bar \SumNode$? 
Basically, we can assume arbitrary normalized weights, which will cause $\bar \SumNode$ to constantly output 1, where, however, a natural choice would be to use uniform weights for $\bar \SumNode$ (maximizing the entropy of the resulting LV model).
Although the choice of weights is not crucial for \emph{evaluating evidence} in the SPN, it plays a role in \emph{MPE inference}, see Section \ref{sec:MPE}.
For now, let us formalize the explicit introduction of LVs, denoted as \emph{augmentation}.

\subsection{Augmentation of SPNs}
\begin{figure}
\centering
 \subfloat[]{
  \includegraphics[scale=0.35]{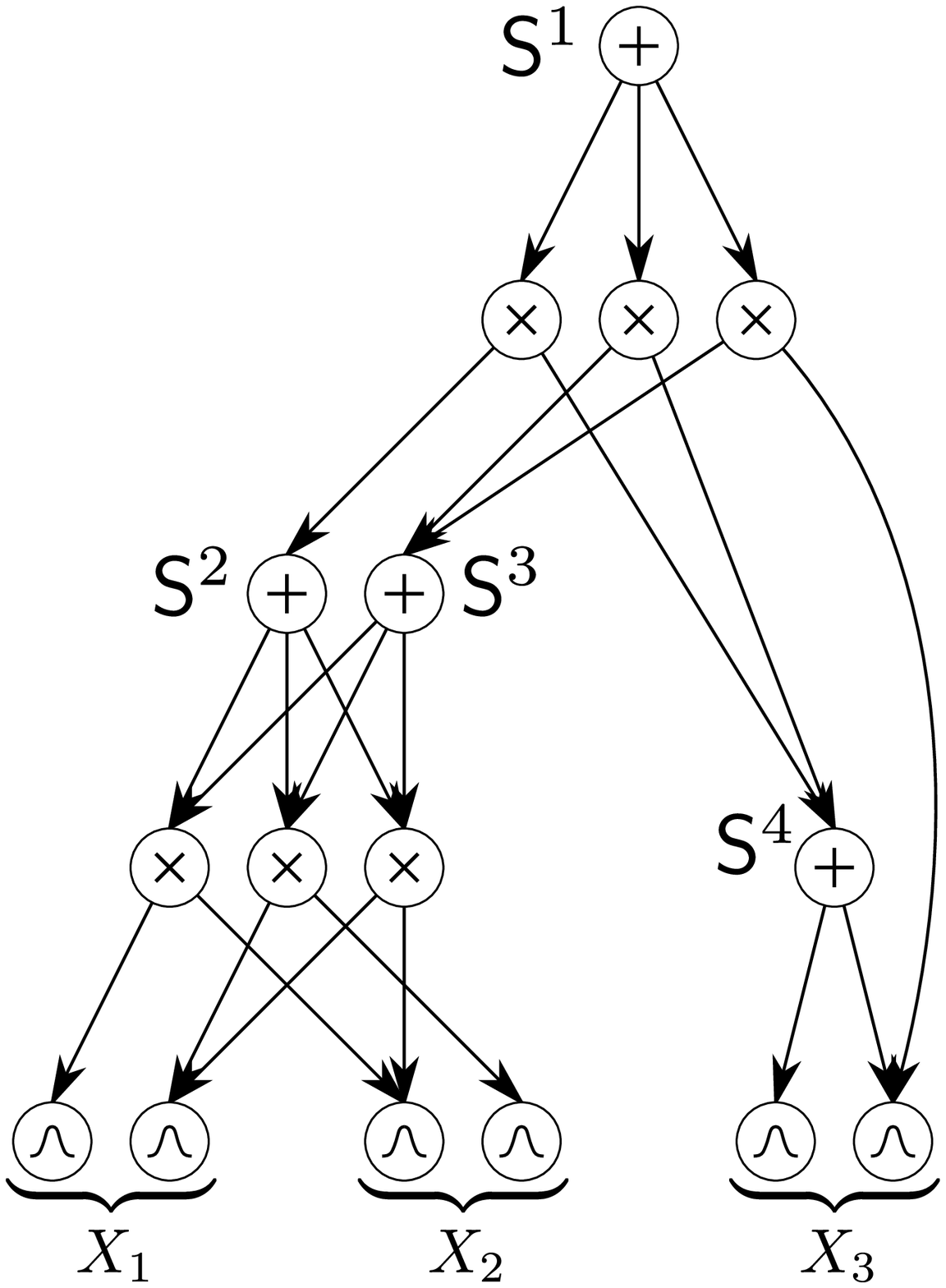}
  \label{fig:exampleaugpre}	
  }\qquad
 \subfloat[]{
  \includegraphics[scale=0.35]{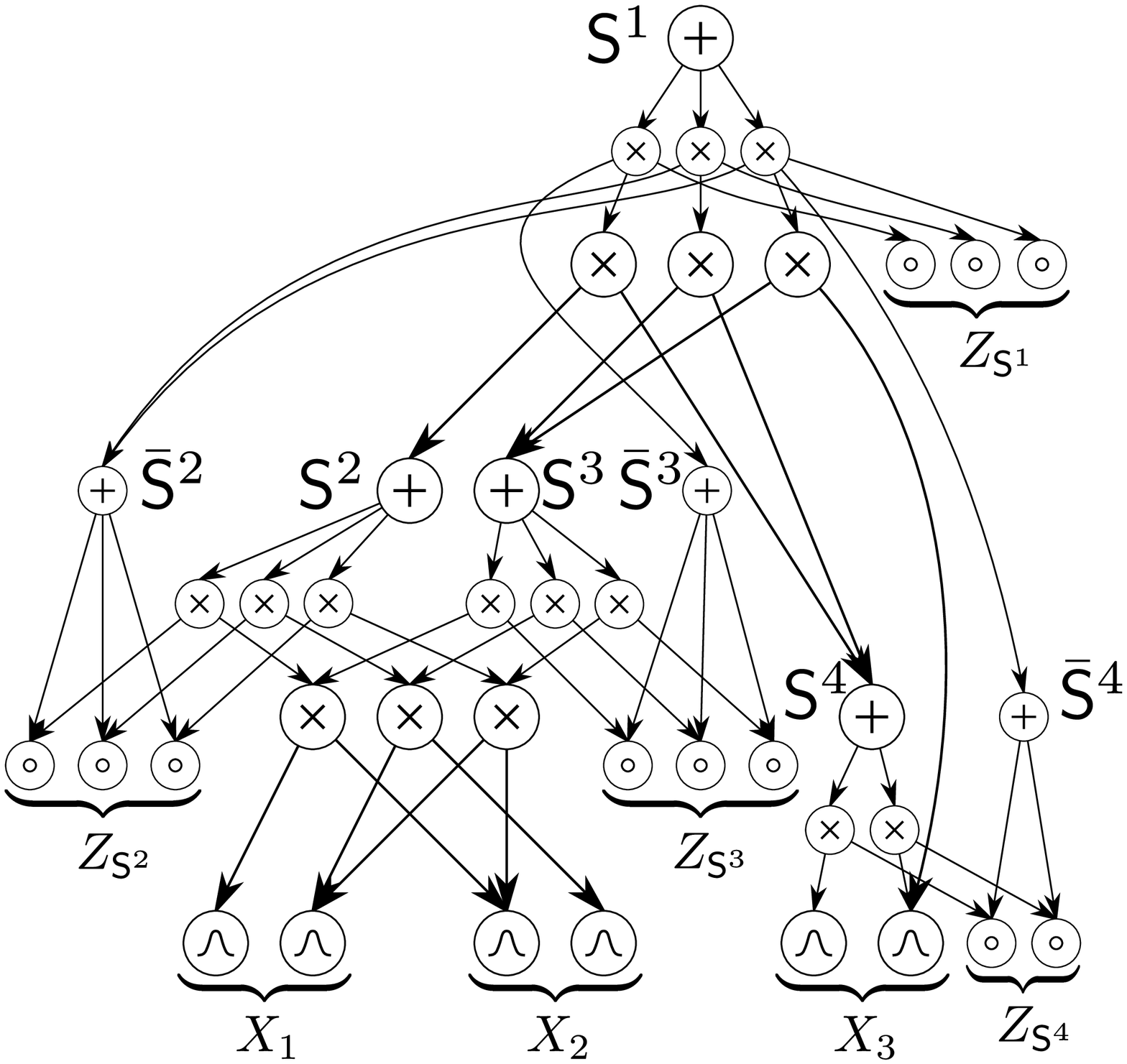}
  \label{fig:exampleaugpost}	
 }
 \caption{Augmentation of an SPN. 
\protect\subref{fig:exampleaugpre}: Example SPN over $\X = \{X_1, X_2, X_3\}$, containing sum nodes $\SumNode^1$, $\SumNode^2$, $\SumNode^3$ and $\SumNode^4$. 
\protect\subref{fig:exampleaugpost}: Augmented SPN, containing IVs corresponding to $Z_{\SumNode^1}$, $Z_{\SumNode^2}$, $Z_{\SumNode^3}$, $Z_{\SumNode^4}$, links and twin sum nodes $\bar{\SumNode}^2$, $\bar{\SumNode}^3$, $\bar{\SumNode}^4$.
For nodes introduced by augmentation, smaller circles are used. 
}
\label{fig:exaugmentSPN}
\end{figure}

Let $\SPN$ be an SPN over $\X$.
For each $\SumNode \in \bm{\SumNode}(\SPN)$ we assume an arbitrary but fixed ordering of its children $\ch(\SumNode) = \{\Child_\SumNode^1,\dots,\Child_\SumNode^{K_\SumNode}\}$, where $K_\SumNode = |\ch(\SumNode)|$.
Let $Z_\SumNode$  be an RV on the same probability space as $\X$, with $\val(Z_\SumNode) = \{1,\dots,K_\SumNode\}$, where state $k$ corresponds to child $\Child^k_\SumNode$.
We call $Z_\SumNode$ the \emph{LV associated with} $\SumNode$.
For sets of sum nodes $\bm{\SumNode}$ we define $\Z_{\bm{\SumNode}} = \{ Z_\SumNode \sbar \SumNode \in \bm{\SumNode} \}$.
To distinguish $\X$ from the LVs, we will refer to the former as \emph{model RVs}.
For node $\Node$, we define the \emph{sum ancestors/descendants} as
\begin{align}
\Sanc(\Node) &:= \anc(\Node) \cap \SumNodes(\SPN),  \\
\Sdesc(\Node) &:= \desc(\Node) \cap \SumNodes(\SPN).
\end{align}
For each sum node $\SumNode$ we define the \emph{conditioning sums} as
\begin{equation}
 \bm{\SumNode}^c(\SumNode) := \{\SumNode^c \in \anc_{\bm{\SumNode}}(\SumNode) \setminus \{\SumNode\} \sbar \exists \Child \in \ch(\SumNode^c) \colon \SumNode \notin \desc(\Child) \}.
\end{equation}
Furthermore, we assume a set of locally normalized \emph{twin-weights} $\bm{\bar{\w}}$, containing a twin-weight $\bar{\w}_{\SumNode,\Child}$ for each weight $\w_{\SumNode,\Child}$ in the SPN.
We are now ready to define the \emph{augmentation} of an SPN. 
\begin{figure}
\begin{algorithmic}[1]
\Procedure{AugmentSPN}{$\SPN$}
\State $\SPN' \leftarrow \SPN$ 
\State $\forall \SumNode \in \bm{\SumNode}(\SPN'),~\forall k \in \{1,\dots,K_\SumNode\}\colon$
\Statex \hskip\algorithmicindent \hskip\algorithmicindent  
let $\w_{\SumNode,k} = \w_{\SumNode,\Child_\SumNode^k}$, $\bar\w_{\SumNode,k} = \bar\w_{\SumNode,\Child_\SumNode^k}$
\For{$\SumNode \in \bm{\SumNode}(\SPN')$}      \label{as:introLinksStart}
\For{$k = 1 \dots K_\SumNode$}
\State Introduce a new product node $\ProdNode_\SumNode^k$ in $\SumNodes(\SPN')$
\State Disconnect $\Child_\SumNode^k$ from $\SumNode$
\State Connect $\Child_\SumNode^k$ as child of $\ProdNode_\SumNode^k$
\State Connect $\ProdNode_\SumNode^k$ as child of $\SumNode$ with weight $\w_{\SumNode,k}$
\EndFor
\EndFor                                        \label{as:introLinksEnd}
\For{$\SumNode \in \bm{\SumNode}(\SPN')$}
\For {$k \in \{1,\dots,K_\SumNode\}$}            \label{as:introIVStart}
\State Connect new IV $\IV{Z_\SumNode}{k}$ as child of $\ProdNode_\SumNode^k$
\EndFor                                           \label{as:introIVEnd}
\If {$\bm{\SumNode}^c(\SumNode) \not= \emptyset$}   
\State Introduce a twin sum node $\bar{\SumNode}$ in $\SPN'$             \label{as:augmentRemedyStart}
\State $\forall k \in \{1,\dots,K_\SumNode\}$: connect $\IV{Z_\SumNode}{k}$ as child of $\bar{\SumNode}$, 
\Statex \hskip\algorithmicindent \hskip\algorithmicindent \hskip\algorithmicindent \hskip\algorithmicindent
and let $\w_{\bar{\SumNode}, \IV{Z_\SumNode}{k}} = \bar{w}_{{\SumNode}, k}$
\For{$\SumNode^c \in \bm{\SumNode}^c(\SumNode)$} 
\For {$k \in \{k ~|~ \SumNode \not\in \desc(\ProdNode^k_{\SumNode^c})\}$}
\State Connect $\bar{\SumNode}$ as child of $\ProdNode^k_{\SumNode^c}$     \label{as:connectTwin}
\EndFor
\EndFor                                                              \label{as:augmentRemedyEnd}
\EndIf                      
\EndFor
\State \Return $\SPN'$
\EndProcedure
\end{algorithmic}
\caption{Pseudo-code for augmentation of an SPN.}
\label{algo:augmentSPN}
\end{figure}
\begin{definition}[Augmentation of SPN]
Let $\SPN$ be an SPN over $\X$, $\bar{\bm{\w}}$ be a set of twin-weights and $\SPN'$ be the result of algorithm \textproc{AugmentSPN}, shown in 
Fig.~\ref{algo:augmentSPN}.
$\SPN'$ is called the \emph{augmented} SPN of $\SPN$, denoted as $\SPN' =: \aug(\SPN)$.
Within the context of $\SPN'$, $\Child_\SumNode^k$ is called the $k^\text{th}$ \emph{former child} of $\SumNode$.
The introduced product node $\ProdNode_\SumNode^k$ is called \emph{link} of $\SumNode$, $\Child^k_\SumNode$ and $\IV{Z_\SumNode}{k}$, respectively.
The sum node $\bar{\SumNode}$, if introduced, is called the \emph{twin} sum node of $\SumNode$.
With respect to $\SPN'$, we denote $\SPN$ as the \emph{original} SPN.
\end{definition}
In steps \ref{as:introLinksStart}--\ref{as:introLinksEnd} of \textproc{AugmentSPN} we introduce the links $\ProdNode_\SumNode^k$ which are interconnected between sum node $\SumNode$ and its $k^\text{th}$ child.
Each link $\ProdNode_\SumNode^k$ has a single parent, namely $\SumNode$, and simply copies the former child $\Child_\SumNode^k$.
In steps \ref{as:introIVStart}--\ref{as:introIVEnd}, we introduce IVs corresponding to the associated LV $Z_\SumNode$, as proposed in \cite{Poon2011}.
As we saw in Fig.~\ref{fig:problemAugmentSPNnaiv} and the discussion above, this can render other sum nodes incomplete.
These sums are clearly the conditioning sums $\bm{\SumNode}^c(\SumNode)$.
Thus, when necessary, we introduce a twin sum node in steps \ref{as:augmentRemedyStart}--\ref{as:augmentRemedyEnd}, to treat this problem.
The following proposition states the soundness of augmentation.
\begin{proposition}
\label{prop:augmentationSound}
Let $\SPN$ be an SPN over $\X$, $\SPN' = \aug(\SPN)$ and $\Z := \Z_{\bm{\SumNode}(\SPN)}$.
Then $\SPN'$ is a complete and decomposable SPN over $\X \cup \Z$ with ${\SPN'}(\X) \equiv \SPN(\X)$.
\end{proposition}
Proposition \ref{prop:augmentationSound} states that the marginal distribution over $\X$ in the augmented SPN is the same distribution 
as represented by the original SPN, while being a \emph{completely specified probabilistic model} over $\X$ \emph{and} $\Z$. 
Thus, augmentation provides a sound way to generalize the LV interpretation from mixture models to more general SPNs. 
An example of augmentation is shown in Fig.~\ref{fig:exaugmentSPN}. 

Note that we understand the augmentation mainly as a \emph{theoretical} tool to establish and work with the LV interpretation in SPNs.
In most cases, it will be neither necessary nor advisable to \emph{explicitly} construct the augmented SPN.

An interesting question is how the sizes of the original SPN and the augmented SPN relate to each other.
A lower bound is $|\SPN'| \in \Omega(|\SPN|)$, holding e.g.~for SPNs with a single sum node.
An asymptotic upper bound is $|\SPN'| \in \mathcal{O}(|\SPN|^2)$.
To see this, note that the introduction of links, IVs and twin sums cause at most a linear increase of the SPN's size.
The number of edges introduced when connecting twins to the links of conditioning sums is bounded by $|\SPN|^2$, 
since the number of twins and links are both bounded by $|\SPN|$.
Therefore, we have $|\SPN'| \in \mathcal{O}(|\SPN|^2)$.
%
%
%
This asymptotic upper bound is indeed achieved by certain types of SPNs:
Consider e.g.~a chain consisting of $K$ sum nodes and $K+1$ distribution nodes.
For $k<K$ the $k^\text{th}$ sum is the parent of the $(k+1)^\text{th}$ sum and the $k^\text{th}$ distribution, and the $K^\text{th}$ sum is the parent of the last two distributions.
For the $k^\text{th}$ sum, all preceding sums are conditioning sums, yielding $k-1$ introduced edges.
In total this gives $\sum_{k=2}^{K} (k-1) = \frac{K\,(K-1)}{2} = \frac{K^2 - K}{2}$ edges, i.e.~in this case $|\SPN'|$ indeed grows quadratically in $|\SPN|$.

\subsection{Conditional Independencies in Augmented SPNs and Probabilistic Interpretation of Sum-Weights}
\begin{figure}
\centering
\includegraphics[width=0.21\textwidth]{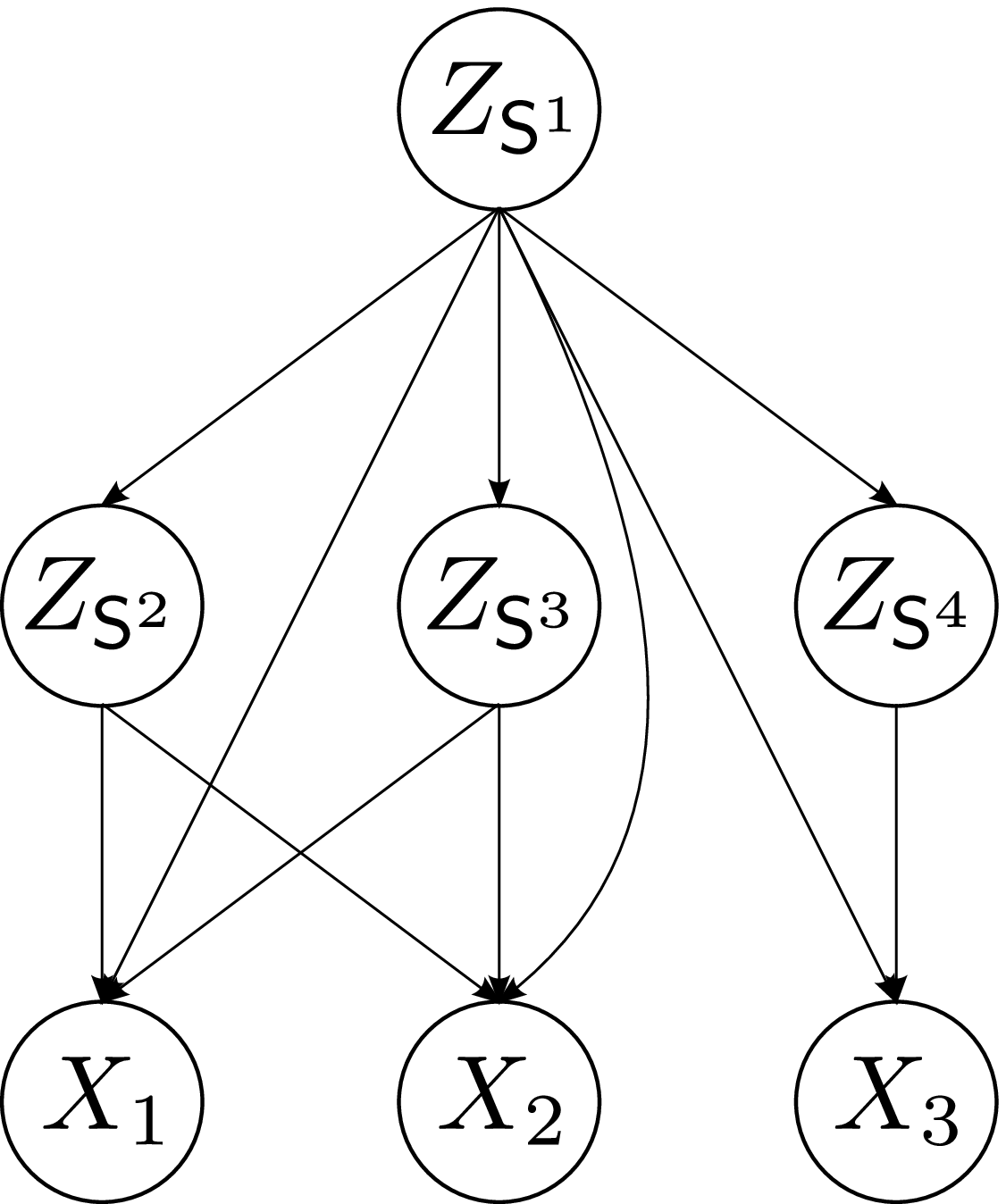}
\label{fig:reaugspn_BN}
\caption{Dependency structure of augmented SPN from Fig.~\ref{fig:exaugmentSPN}, represented as BN.
}
\label{fig:dependencystructBN}
\end{figure}

It is helpful to introduce the notion of configured SPNs, which takes a similar role as conditioning in the literature on DNNFs \cite{Darwiche1999, Darwiche2001, Darwiche2002}.
%
%
\begin{definition}[Configured SPN]
Let $\SPN$ be an SPN over $\X$, $\Y \subseteq \Z_{\SumNodes(\SPN)}$ and $\y \in \val(\Y)$.
The \emph{configured SPN} $\SPN^\y$ is obtained by deleting the IVs $\IV{Y}{y}$ and their corresponding link for each $Y \in \Y$, $y \not= \subv{\y}{Y}$ from 
$\aug(\SPN)$, and further deleting all nodes which are rendered unreachable from the root.
\end{definition}
Intuitively, the configured SPN isolates the computational structure selected by $\y$. 
All sum edges which "survive" in the configured SPN are equipped with the same weights as in the augmented SPN. 
Therefore, a configured SPN is in general not locally normalized. 
We note the following properties of configured SPNs.
\begin{proposition}
\label{prop:confspn}
Let $\SPN$ be an SPN over $\X$, $\Y \subseteq \Z_{\SumNodes(\SPN)}$ and $\Z = \Z_{\SumNodes(\SPN)} \setminus \Y$.
Let $\y \in \val(\Y)$ and let $\SPN' = \aug(\SPN)$.  
It holds that 
\begin{enumerate}
 \item Each node in $\SPN^\y$ has the same scope as its corresponding node in $\SPN'$.
 \item $\SPN^\y$ is a complete and decomposable SPN over $\X \cup \Y \cup \Z$.
 \item For any node $\Node$ in $\SPN^\y$ with $\scope(\Node) \cap \Y = \emptyset$, we have that $\SPN^\y_\Node = \SPN'_\Node$.
 \item For $\y' \in \val(\Y)$ it holds that 
\begin{equation}
 \SPN^\y(\X,\Z,\y') = 
\begin{cases}
 \SPN'(\X,\Z,\y') & \text{if } \y' = \y \\
0                 & \text{otherwise}
\end{cases}
\end{equation}
\end{enumerate}
\end{proposition}
The next theorem shows certain conditional independencies in the augmented SPN. 
For ease of discussion, we make the following definitions.
\begin{definition}
Let $\SumNode$ be a sum node in an SPN and $Z_\SumNode$ its associated LV.
All other RVs (model RVs and LVs) are divided into three sets:
\begin{itemize}
\item
\emph{Parents} $\Z_p$, which are all LVs "above" $\SumNode$, i.e.~$\Z_p = \Z_{\Sanc(\SumNode)} \setminus Z_\SumNode$.
\item 
\emph{Children} $\Y_c$, which are all model RVs and LVs "below" $\SumNode$, i.e.~$\Y_c = \scope(\SumNode) \cup \Z_{\Sdesc(\SumNode)} \setminus Z_\SumNode$.
\item
\emph{Non-descendants} $\Y_n$, which are the remaining RVs, i.e.~$\Y_n = (\X \cup \Z_{\SumNodes(\SPN)}) \setminus (\Z_p \cup \Y_c \cup Z_\SumNode)$.
\end{itemize}
\end{definition}
We will show that the \emph{parents}, \emph{children} and \emph{non-descendants}
play the likewise role as for independencies in BNs \cite{Pearl1988, Koller2009},
i.e. $Z_\SumNode$ \emph{is independent of} $\Y_n$ \emph{given} $\Z_p$. 
We will further show that the sum-weights of $\SumNode$ are the conditional distribution
of $Z_\SumNode$, conditioned on the event that "$\Z_p$ select a path to $\SumNode$". 
One problem in the original LV interpretation \cite{Poon2011} was, that no conditional distribution of $Z_\SumNode$ was specified for
the complementary event.
Here, we will show that the twin-weights are precisely this conditional distribution. 
This requires that the event ``$\Z_p$ select a path to the twin $\bar\SumNode$'' is indeed the complementary event to ``$\Z_p$ select a path to $\SumNode$''.
This is shown in following lemma.
\begin{lemma}
\label{lem:singlepath}
Let $\SPN$ be an SPN over $\X$, let $\SumNode$ be a sum node in $\SPN$ and $\Z_p$ be the parents of $Z_\SumNode$.
For any $\z \in \val(\Z_p)$, the configured SPN $\SPN^{\z}$ contains either $\SumNode$ or its twin $\bar{\SumNode}$, but not both.
\end{lemma}
We are now ready to state the our theorem concerning conditional independencies in augmented SPNs.
\begin{theorem}
\label{theo:conditionalIndependence}
Let $\SPN$ be an SPN over $\X$ and $\SPN' = \aug(\SPN)$.
Let $\SumNode$ be an arbitrary sum in $\SPN$ and $\w_k = \w_{\SumNode,\Child^k_\SumNode}$, $\bar\w_k = \bar\w_{\SumNode,\Child^k_\SumNode}$, 
$k=1,\dots,K_\SumNode$.
With respect to $\SumNode$, let $\Z_p$ be the parents, $\Y_c$ be the children and $\Y_n$ be the 
non-descendants, respectively.
Then there exists a two-partition of $\val(\Z_p)$, i.e.~$\ZS, \bar\ZS \colon \ZS \cup \bar\ZS = \val(\Z_p)$, $\ZS \cap \bar\ZS = \emptyset$, such that
\begin{align}
\forall \z \in \ZS \colon \SPN'(Z_\SumNode = k, \Y_n, \z) & = \w_k \SPN'(\Y_n,\z), and   \label{eq:CIaugSum}\\
\forall \z \in \bar \ZS \colon \SPN'(Z_\SumNode = k, \Y_n, \z) & = \bar\w_k \SPN'(\Y_n,\z).  \label{eq:CIaugTwinSum}
\end{align}
\end{theorem}
From Theorem~\ref{theo:conditionalIndependence} it follows that the weights and twin-weights of a sum node $\SumNode$ 
can be interpreted as \emph{conditional probability tables} (CPTs) of $Z_\SumNode$, conditioned on $\Z_p$ and that 
$Z_\SumNode$ is conditionally independent of $\Y_n$ given $\Z_p$, i.e.
\begin{equation}
\label{eq:weightsAsCondProbs}
\SPN'(Z_\SumNode = k \cbar \Y_n, \z) = \SPN'(Z_\SumNode = k \cbar \z) = 
\begin{cases}
\w_k & \text{if } \z \in \ZS \\
\bar\w_k & \text{if } \z \in \bar \ZS.
\end{cases}
\end{equation}
Using this result, we can define a BN representing the
augmented SPN as follows: For each sum node $\SumNode$, connect
$\Z_p$ as parents of $Z_\SumNode$, and all RVs $\scope(\SumNode)$ as children of $Z_\SumNode$. 
By doing this for each LV, we obtain our BN representation of the augmented SPN, serving as a useful tool 
to understand SPNs in the context of probabilistic graphical models. 
An example of the BN interpretation is shown in Fig.~\ref{fig:dependencystructBN}. 

Note that the BN representation by Zhao et al. \cite{Zhao2015} can be recovered from the BN representation of augmented SPNs.
They proposed a BN representation of SPNs using a bipartite structure, where an LV is a parent of a model RV if it is 
contained in the scope of the corresponding sum node.
The model RVs and LVs are unconnected among each other, respectively. 
When we constrain the twin-weights to be equal to the sum-weights, we can see in \eqref{eq:weightsAsCondProbs} that $Z_\SumNode$ becomes independent of $\Z_p$.
This special choice of twin weights effectively removes all edges between LVs, recovering the BN structure in \cite{Zhao2015}.
In the next section, we use the augmented SPN and the BN interpretation to derive the EM algorithm for SPNs.

\section{EM Algorithm}
\label{sec:EM}
The EM algorithm is a general scheme for maximum likelihood
learning, when for some RVs complete evidence is missing \cite{Dempster1977, Ghahramani1994}. 
Thus, augmented SPNs are amenable for EM due to the LVs associated with sum nodes. 
Moreover, the twin-weights can be kept fixed, so that EM applied to augmented SPNs actually optimizes the weights of the original SPN. 
This approach was already pointed out in \cite{Poon2011}, where it was suggested that for evidence $\e$ and for any LV $Z_\SumNode$, the marginal posteriors should be given as $p(Z_\SumNode = k \cbar \e) \propto \w_{\SumNode, \Child^k_\SumNode} \frac{\partial \SPN(\e)}{\partial \SumNode(\e)}$, which should be used for EM updates.
These updates, however, cannot be the correct ones, as they actually \emph{leave the weights unchanged}.
Here, using augmented SPNs, we formally derive the standard EM updates for sum-weights and the input distributions, when they are chosen from an exponential family.

\subsection{Updates for Weights}
\label{sec:MPE_weights}
\begin{figure}
\centering
 \subfloat[]{
  \includegraphics[scale=0.35]{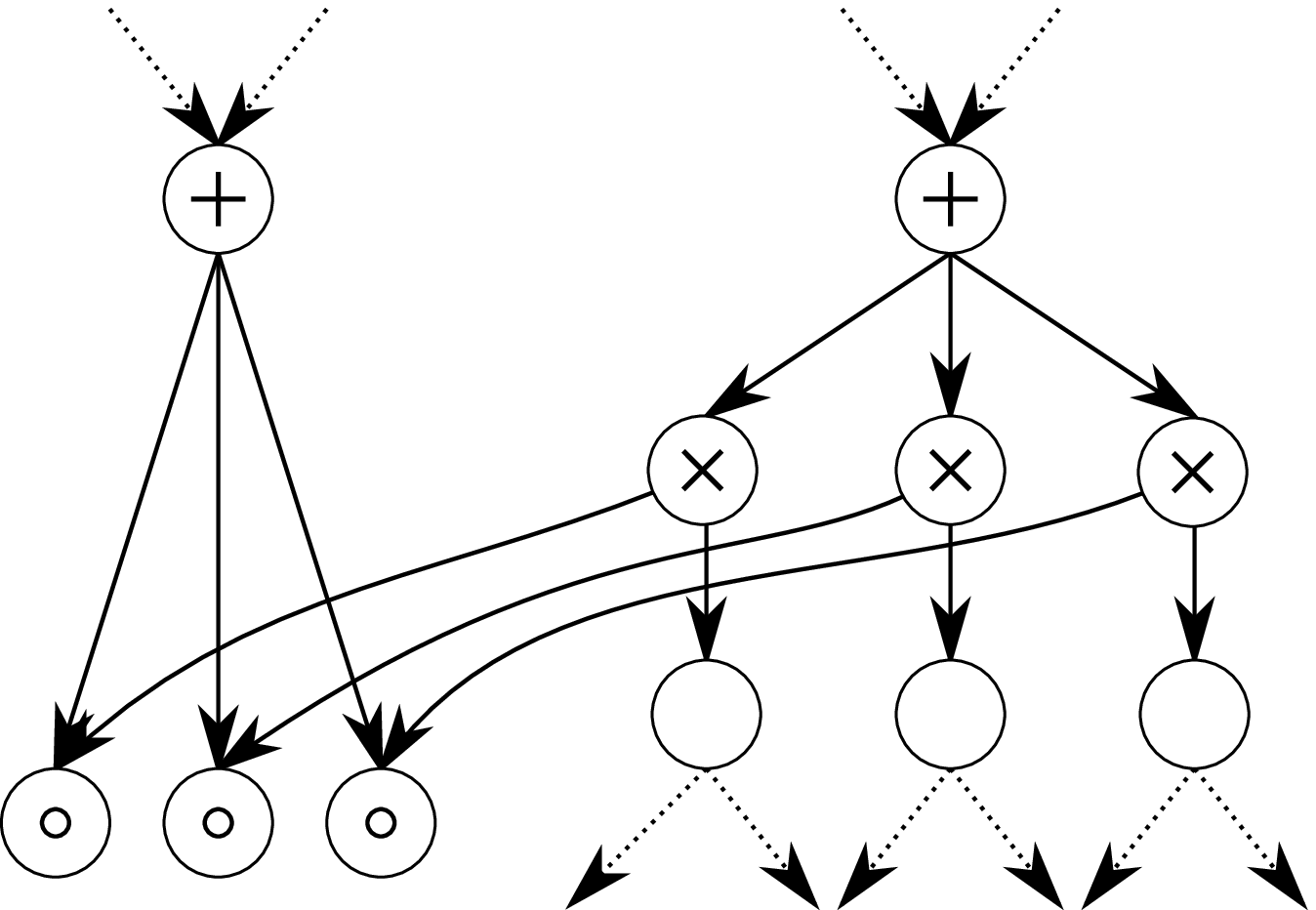}
  \label{fig:augSPN_switchpar1}
  \put(-55,72){$\SumNode$}
  \put(-130,72){$\bar{\SumNode}$}
  \put(-150,5){$\underbrace{~~~~~~~~~~~~~~~~~~}_{\IV{Z_{\SumNode}}{1} \, \IV{Z_{\SumNode}}{2} \, \IV{Z_{\SumNode}}{3}}$}
  }\qquad
 \subfloat[]{
  \includegraphics[scale=0.35]{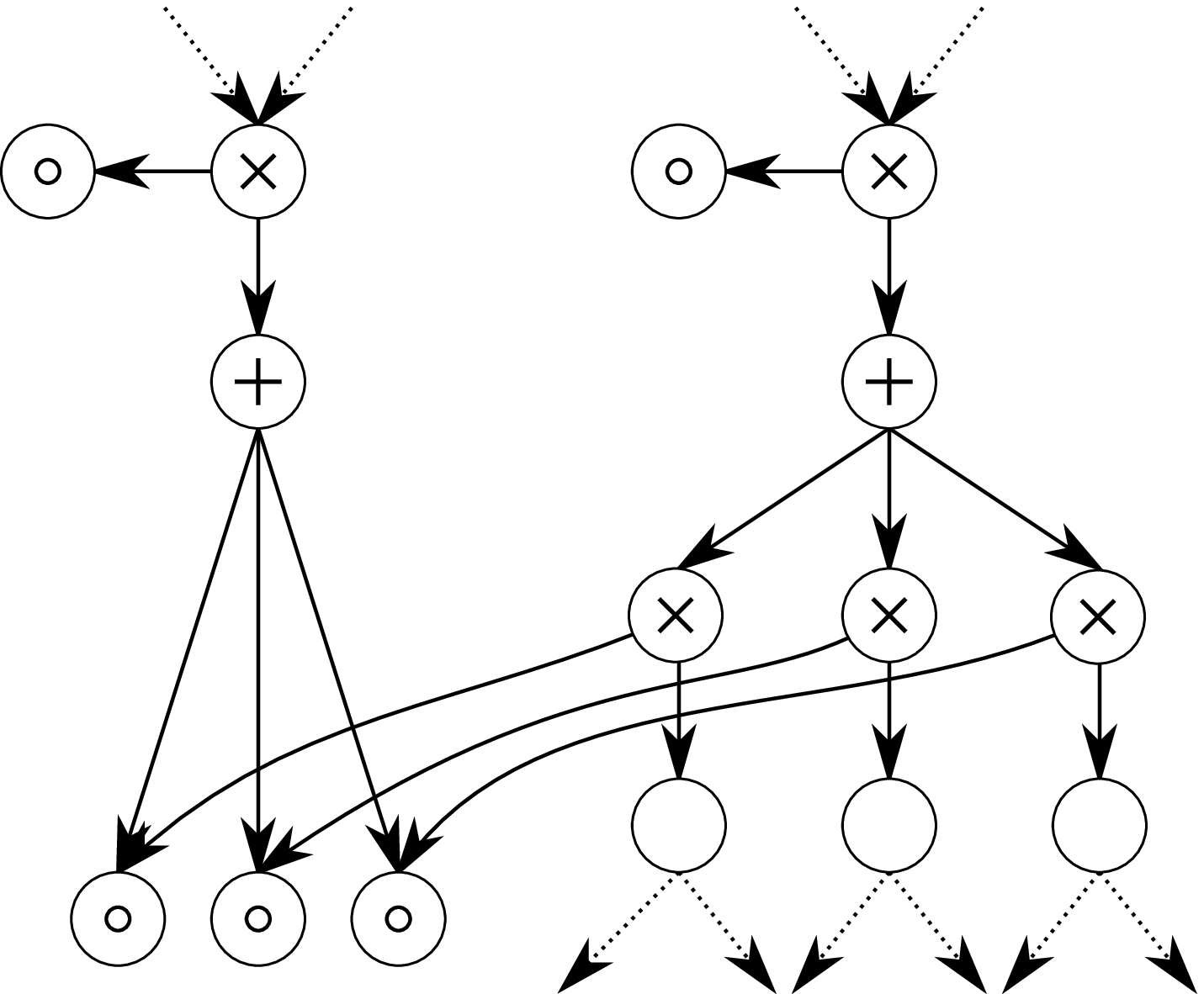}
   \label{fig:augSPN_switchpar2}
	 \put(-55,72){$\SumNode$}
   \put(-130,72){$\bar{\SumNode}$}
   \put(-150,5){$\underbrace{~~~~~~~~~~~~~~~~~~}_{\IV{Z_{\SumNode}}{1} \, \IV{Z_{\SumNode}}{2} \, \IV{Z_{\SumNode}}{3}}$}
   \put(-78,87){$\IV{Y_\SumNode}{y_{\SumNode}}$}
   \put(-153,87){$\IV{Y_\SumNode}{y_{\bar{\SumNode}}}$}
   \put(-30,97){$\ProdNode$}
}
\caption{Explicitly introducing a switching parent $Y_\SumNode$ in an augmented SPN. 
\protect\subref{fig:augSPN_switchpar1}: Part of an augmented SPN containing a sum node with three children and its twin. 
\protect\subref{fig:augSPN_switchpar2}: Explicitly introduced switching parent $Y_\SumNode$ using IVs $\IV{Y_\SumNode}{y_{\SumNode}}$ and $\IV{Y_\SumNode}{y_{\bar{\SumNode}}}$.}
\label{fig:augmentSPN_switchingparent}
\end{figure}

Assume a dataset $\mathcal{D} = \{\e^{(1)}, \dots, \e^{(L)}\}$ of $L$ i.i.d.~samples, where each $\e^{(l)}$
is any combination of complete and partial evidence for the model RVs $\X$, cf.~Section~\ref{sec:backgroundNotation}.
Let $\Z = \Z_{\SumNodes(\SPN)}$ be the set of all LVs and consider an arbitrary sum node $\SumNode$.
Eq.~\eqref{eq:weightsAsCondProbs} shows that the weights can be interpreted as conditional probabilities
in our BN interpretation, where
\begin{equation}
\SPN'(Z_\SumNode = k \cbar \Z_p = \z) = 
\begin{cases}
\w_k & \text{if } \z \in \ZS \\
\bar\w_k & \text{if } \z \in \bar \ZS.
\end{cases}
\end{equation}
As mentioned above, the twin-weights $\bar\w_k$ are kept fixed. 
Using the well-known EM-updates in BNs over discrete RVs \cite{Koller2009, Dempster1977}, the updates for sum-weight $\w_k$ are given by summing over the expected statistics
\begin{equation}
\label{eq:expectedStatisticsWeights}
\SPN'(Z_\SumNode = k, \Z_p \in \ZS \cbar \e^{(l)}),
\end{equation}
followed by renormalization.
We make the event $\Z_p \in \ZS$ explicit, by introducing a \emph{switching parent} $Y_\SumNode$ of $Z_\SumNode$:
When the twin sum of $\SumNode$ exists, $Y_\SumNode$ assumes the two states $\val(Y_\SumNode)=\{y_\SumNode, y_{\bar \SumNode}\}$, where 
$Y_\SumNode = y_\SumNode \Leftrightarrow \Z_p \in \ZS$ and 
$Y_\SumNode = y_{\bar\SumNode} \Leftrightarrow \Z_p \in \bar\ZS$.
When the twin sum does not exist, $Y_\SumNode$ just takes the single value $\val(Y_\SumNode)=\{y_\SumNode\}$. 
Clearly, when observed, $Y_\SumNode$ renders $Z_\SumNode$ independent from $\Z_p$. 
The switching parent can be explicitly introduced in the augmented SPN, as depicted in Fig.~\ref{fig:augmentSPN_switchingparent}. 
Here we simply introduce two new IVs $\IV{Y_\SumNode}{y_\SumNode}$ and $\IV{Y_\SumNode}{y_{\bar\SumNode}}$, 
which switch on/off the output of $\SumNode$ and $\bar\SumNode$, respectively. 
It is easy to see that when these IV are constantly set to 1, i.e. when $Y_\SumNode$ is marginalized, 
the augmented SPN performs exactly the same computations as before. 
It is furthermore easy to see that completeness and decomposability of the augmented SPN are maintained
when the switching parent is introduced. 
Using the switching parent, the required expected statistics \eqref{eq:expectedStatisticsWeights} translate to
\begin{equation}
\label{eq:expectedStatisticsWeights2}
\SPN'(Z_\SumNode = k, Y_\SumNode = y_\SumNode \cbar \e^{(l)}).
\end{equation}
To compute \eqref{eq:expectedStatisticsWeights2}, we use the differential approach, \cite{Darwiche2003, Peharz2015, Peharz2015b}, cf.~also Section~\ref{sec:backgroundNotation}.
First note that
\begin{equation}
\SPN'(Z_\SumNode = k, Y_\SumNode = y_\SumNode, \e^{(l)}) = 
\frac
{\partial^2 \SPN'(\e^{(l)})}
{\partial \IV{Y_\SumNode}{y_\SumNode} \partial \IV{Z_\SumNode}{k}}.
\end{equation}
The first derivative is given as
\begin{align}
\frac{\partial \SPN'(\e^{(l)})}{\partial \IV{Y_\SumNode}{y_\SumNode}} 
&= \frac{\partial \SPN'(\e^{(l)})}{\partial \ProdNode} \, \SumNode(\e^{(l)}) \\
&=  \frac{\partial \SPN'(\e^{(l)})}{\partial \ProdNode} \, \sum_{k=1}^{K_\SumNode} \IV{Z_\SumNode}{k} \, \w_k \, \Child_\SumNode^k(\e^{(l)}), 
\label{eq:firstDerivativeWeights}
\end{align}
where $\ProdNode$ is the common product parent of $\SumNode$ and $ \IV{Y_\SumNode}{y_\SumNode}$ in the augmented SPN (see Fig.~\ref{fig:augSPN_switchpar2}).
Differentiating \eqref{eq:firstDerivativeWeights} after $\IV{Z_\SumNode}{k}$ yields the second derivative 
\begin{equation}
\frac
{\partial^2 \SPN'(\e^{(l)})}
{\partial \IV{Y_\SumNode}{y_\SumNode} \partial \IV{Z_\SumNode}{k}}
=
\frac{\partial \SPN'(\e^{(l)})}{\partial \ProdNode} \, \w_k \, \Child_\SumNode^k(\e^{(l)}),
\end{equation}
delivering the required posteriors
\begin{equation}
\SPN'(Z_\SumNode = k, Y_\SumNode = y_\SumNode \cbar \e^{(l)}) = 
\frac{1}{\SPN'(\e^{(l)})}
\frac{\partial \SPN'(\e^{(l)})}{\partial \ProdNode} \, \w_k \, \Child_\SumNode^k(\e^{(l)}).
\label{eq:EMposteriorsWeight}
\end{equation}
We do not want to construct the augmented SPN explicitly, so we express \eqref{eq:EMposteriorsWeight} in terms of the original SPN.
Since all LVs are marginalized, it holds that $\SPN'(\e^{(l)}) = \SPN(\e^{(l)})$ and $\frac{\partial \SPN'(\e^{(l)})}{\partial \ProdNode} = \frac{\partial \SPN(\e^{(l)})}{\partial \SumNode}$, yielding
\begin{equation}
\SPN'(Z_\SumNode = k, Y_\SumNode = y_\SumNode \cbar \e^{(l)}) = 
\frac{1}{\SPN(\e^{(l)})}
\frac{\partial \SPN(\e^{(l)})}{\partial \SumNode} \, \w_k \, \Child_\SumNode^k(\e^{(l)}),
\label{eq:EMposteriorsWeight2}
\end{equation}
delivering the required statistics for updating the sum-weights.
We now turn to the updates of the input distributions.

\subsection{Updates for Input Distributions}
\label{sec:MPE_dist}
For simplicity, we derive updates for univariate input distributions,
i.e. for all distributions $\DistNode_\Y$ we have $|\scope(\DistNode_\Y)| = 1$.
Similar updates can rather easily be derived also for multivariate
input distributions. 
In \cite{Peharz2015}, the so-called distribution selectors (DSs) were introduced 
to derive the differential approach for generalized SPNs. 
Similar as the switching parents for (twin) sum nodes, the DSs are RVs which render
the respective model RVs independent from the remaining RVs. 
More formally, for each $X \in \X$, let $\DistNodes_X$ be the set of
all input distributions which have scope $\{X\}$. 
Assume an arbitrary but fixed ordering of $\DistNodes_X$ and let $[\DistNode_X]$ be the index
of $\DistNode_X$ in this ordering. 
Let the DS $W_X$ be a discrete RV with $|\DistNodes_X|$ states. 
The so-called gated SPN $\SPN^g$ is obtained by replacing each distribution by the product node
\begin{equation}   \label{eq:introduceGate}
\DistNode_X \rightarrow \DistNode_X \times \IV{W_X}{[\DistNode_X]}.
\end{equation}
The introduced product is denoted as gate. 
As shown in \cite{Peharz2015}, $X$ is rendered independent from all other RVs in the SPN
when conditioned on $W_X$. 
Moreover, $\DistNode_X$ is the conditional distribution of $X$ given 
$W_X = [\DistNode_X]$. 
Therefore, each $X$ and its DS $W_X$ can be incorporated as a two RV family in our BN
interpretation. 
When each input distribution $\DistNode_X$ is chosen from an exponential family with natural parameters 
$\theta_{\DistNode_X}$, the M-step is given by the expected sufficient statistics
\begin{equation}   \label{eq:MstepDistnodes}
\theta_{\DistNode_X} \leftarrow 
\frac
{\sum_l \SPN^g(W_X = k \cbar \e^{(l)}) \int \DistNode_X(x \cbar \e^{(l)}) \theta_{\DistNode_X}(x) \mathrm{d}x}
{\sum_l \SPN^g(W_X = k \cbar \e^{(l)})},
\end{equation}
where $k=[\DistNode_X]$.
When $\e^{(l)}$ contains complete evidence $x'$ for $X$, then the integral 
$\int \DistNode_X(x \cbar \e^{(l)}) \theta_{\DistNode_X}(x) \mathrm{d}x$
reduces to $\theta_{\DistNode_X}(x')$.
When $\e^{(l)}$ contains partial evidence $\xs$, then 
\begin{equation} \label{eq:integralDist}
\int \DistNode_X(x \cbar \e^{(l)}) \theta_{\DistNode_X}(x) \mathrm{d}x =
\frac
{ \int_\xs \DistNode_X(x) \theta_{\DistNode_X}(x) \mathrm{d}x }
{ \int_\xs \DistNode_X(x) \mathrm{d}x }.
\end{equation}
Depending on $X$ and the the type of $\DistNode_X$, evaluating \eqref{eq:integralDist} can be more
or less demanding. 
A simple but practical case is when $\DistNode_X$ is Gaussian and $\xs$ is some interval, permitting a
closed form solution for integrating the Gaussian's statistics $\theta(x) = (x,x^2)$, using truncated Gaussians \cite{Johnson1994}.

To obtain the posteriors $\SPN^g(W_X = k \cbar \e^{(l)})$ required in \eqref{eq:MstepDistnodes}, we again use the differential approach.
Note that
\begin{equation}
\SPN^g(W_X = k, \e^{(l)}) = 
\frac
{\partial \SPN^g(\e^{(l)})}
{\partial \IV{W_X}{k}} =
\frac
{\partial \SPN^g(\e^{(l)})}
{\partial \ProdNode}
\DistNode_X(\e^{(l)}),
\end{equation}
where $k=[\DistNode_X]$ and $\ProdNode$ is the gate of $\DistNode_X$, cf.~\eqref{eq:introduceGate}.
If we do not want to construct the gated SPN explicitly, we can use the identity
$
\frac{\partial \SPN^g(\e^{(l)})}{\partial \ProdNode}
=
\frac{\partial \SPN(\e^{(l)})}{\partial \DistNode_X}
$.
Thus the required posteriors are given as
\begin{equation}
\SPN^g(W_X = k \cbar \e^{(l)}) = 
\frac{1}{\SPN(\e^{(l)})}
\frac{\partial \SPN(\e^{(l)})}{\partial \DistNode_X} \DistNode_X(\e^{(l)}).
\end{equation}
The EM algorithm for SPNs, both for sum-weights and input distributions, is summarized in Fig.~\ref{algo:EM}.
In Section \ref{sec:experimentsEM} we empirically verify our derivation of EM and show that standard EM successfully 
trains SPNs when a suitable structure is at hand.

Note that recently Zhao and Poupart~\cite{Zhao2016} derived a concave-convex procedure (CCCP) which yield the same sum-weight updates as the EM algorithm presented here and in \cite{Peharz2015b}.
This result is surprising, as EM and CCCP are rather different approaches in general.

\begin{figure}
\begin{algorithmic}[1]
\Procedure{Expectation-Maximization}{\SPN}
\State Initialize $\bm{\w}$ and input distributions
\While {not converged} 
\State $\forall \SumNode \in \bm{\SumNode}(\SPN), \forall \Child \in \ch(\SumNode)\colon n_{\SumNode,\Child} \leftarrow 0$
\State $\forall X \in \X$, $\forall \DistNode_X \in \DistNodes_X \colon \theta_{\DistNode_X} \leftarrow 0$, $n_{\DistNode_X} \leftarrow 0$
\For{$\sampC = 1 \dots \numSamples$}
\State Input $\mathbf{e}^{(l)}$ to $\SPN$
\State Evaluate $\SPN$ (upward-pass)
\State Backprop $\SPN$ (backward-pass)
\For{$\SumNode \in \bm{\SumNode}(\SPN), \Child \in \ch(\SumNode)$}
\State  $n_{\SumNode,\Child} \leftarrow n_{\SumNode,\Child} + 
\frac{1}{\SPN} \,
\frac{\partial \SPN}{\partial \SumNode} \, 
\Child  \, 
\w_{\SumNode, \Child}$
\EndFor
\For{$X \in \X$, $\DistNode_X \in \DistNodes_X$}
\If {$\e^{(l)}$ is complete w.r.t.~$X$}
\State $x \leftarrow$ complete evidence for $X$
\State $\theta \leftarrow \theta(x)$
\Else
\State $\xs \leftarrow $ partial evidence for $X$
\State $\theta \leftarrow \frac{\int_\xs \DistNode_X(x) \theta(x) \mathrm{d}x}{\int_\xs \DistNode_X(x) \mathrm{d}x}$
\EndIf
\State $p \leftarrow \frac{1}{\SPN}\frac{\partial \SPN}{\partial \DistNode_X} \DistNode_X$
\State $\theta_{\DistNode_X} \leftarrow \theta_{\DistNode_X} + p\,\theta$ 
\State $n_{\DistNode_X} \leftarrow n_{\DistNode_X} + p$ 
\EndFor
\EndFor
\State $\forall \SumNode \in \bm{\SumNode}(\SPN), \forall \Child \in \ch(\SumNode) \colon$ 
$\w_{\SumNode, \Child} \leftarrow \frac{n_{\SumNode, \Child}}{\sum_{\Child' \in \ch(\SumNode)} n_{\SumNode, \Child'}}$
\State $\forall X \in \X, \forall \DistNode_X \in \DistNodes_X \colon$ 
set parameters to $\frac{\theta_{\DistNode_X}}{n_{\DistNode_X}}$
\EndWhile 
\State \Return $\SPN$
\EndProcedure
\end{algorithmic}
\caption{Pseudo-code for EM algorithm in SPNs.}
\label{algo:EM}
\end{figure}

\section{Most Probable Explanation}
\label{sec:MPE}
In \cite{Poon2011,Peharz2013,Peharz2014}, SPNs were applied for reconstructing data using MPE inference.
Given some distribution $\p$ over $\X$ and evidence $\e$, MPE can be formalized as finding
$\underset{\x \in \e}{\arg\max} ~ \p(\x),$
where we assume that $\p$ actually has a maximum in $\e$.
MPE is a special case of MAP, defined as finding
$ \underset{\y \in \subv{\e}{\Y}}{\arg\max} ~ \int_{\subv{\e}{\Z}} \p(\y,\z) \, \mathrm{d}\z,$
for some two-partition of $\X$, i.e.~$\X =\Y \cup \Z, \Y \cap \Z = \emptyset$.
Both MPE and MAP are generally NP-hard in BNs \cite{Bodlaender2002, Park2004, Kwisthout2011}, and MAP is inherently harder than MPE
\cite{Park2004, Kwisthout2011}. 
Using the result in \cite{deCampos2011}, it follows that MAP inference is NP-hard also in SPNs. 
In particular, Theorem 5 in \cite{deCampos2011} shows that the decision version of MAP is NP-complete for a Naive Bayes model, when the class variable is marginalized. 
Naive Bayes is represented by the augmentation of an SPN with a single sum node, the LV representing the class variable.
Therefore, MAP in SPNs is generally NP-hard.
Since MAP in the augmented SPN representing the Naive Bayes model corresponds to MPE inference in the original SPN, i.e.~a mixture model, it follows that also MPE inference is generally NP-hard in SPNs.
A proof tailored to SPNs can be found in \mbox{\cite{Peharz2015b}}.

However, when considering the the sub-class of \emph{selective} SPNs (cf.~Section~\ref{sec:backgroundNotation} and \cite{Peharz2014b}), an MPE solution can be obtained using a Viterbi-style backtracking algorithm in \emph{max-product networks}.
\begin{definition}[Max-Product Network]
Let $\SPN$ be an SPN over $\X$. 
We define the \emph{max-product network} (MPN) $\hat \SPN$, by replacing each distribution node $\DistNode$ by a \emph{maximizing} distribution node
\begin{equation}
\hat{\DistNode} \colon \HC_{\scope(\DistNode)} \mapsto [0,\infty], \hat{\DistNode}(\YS) := \underset{\y \in \YS}{\max} \, \DistNode(\y),
\end{equation}
and each sum node $\SumNode$ by a \emph{max node} 
\begin{equation}
\hat \SumNode := \underset{\hat \Child \in \ch(\hat \SumNode)}{\max} \w_{\hat \SumNode, \hat \Child} \hat \Child.
\end{equation}
A product node $\ProdNode$ in $\SPN$ corresponds to a product node $\hat\ProdNode$ in $\hat \SPN$.
\end{definition}

\begin{theorem}
\label{theo:MPEaug}
Let $\SPN$ be a selective SPN over $\X$ and let $\hat{\SPN}$ the corresponding MPN.
Let $\Node$ be some node in $\SPN$ and $\hat{\Node}$ its corresponding node in $\hat{\SPN}$.
Then, for every $\XS \in \HC_{\scope(\Node)}$ we have $\hat{\Node}(\XS) = \underset{\x \in \XS}{\max} ~ \Node(\x)$.
\end{theorem}
Theorem \ref{theo:MPEaug} shows that the MPN maximizes the probability in its corresponding selective SPN.
The proof (see appendix) also shows how to actually \emph{find} a maximizing assignment. 
For a product, a maximizing assignment is given by combining the maximizing assignments of its children.
For a sum, a maximizing assignment is given by the maximizing assignment of a single child, whose weighted maximum is maximal among all children.
Here the children's maxima are readily given by the upwards pass in the MPN.
Thus, finding a maximizing assignment of any node in an selective SPN recursively reduces to finding maximizing assignments for the children of this node;
this can be accomplished by a Viterbi-like backtracking procedure.
This algorithm, denoted as \textproc{MPESelective}, is shown in Fig.~\ref{algo:MPESelective}. 
Here $Q$ denotes a queue of nodes, where $Q \curvearrowleft \Node$ and $\Node \curvearrowleft Q$ denote the en-queue and de-queue operations, respectively.
Note that {Theorem \ref{theo:MPEaug}} has already been derived for a special case, namely for arithmetic circuits representing network polynomials of BNs over discrete RVs \cite{Darwiche2014}.

A direct corollary of Theorem~\ref{theo:MPEaug} is that MPE inference is tractable in augmented SPNs, since augmented SPNs are \emph{selective SPNs over $\X$ and $\Z$}.
This can easily be seen in \textproc{AugmentSPN}, as for any $\z$ and any sum $\SumNode$, exactly one IV of $Z_\SumNode$ is set to $1$, causing that at most one child of $\SumNode$ (or $\bar \SumNode$) can be non-zero.
Therefore, we can use \textproc{MPESelective} in augmented SPNs, in order to find an MPE solution over \emph{both} model RVs and LVs.
Note that an MPE solution for the augmented SPN does in general \emph{not} correspond to an MPE solution for the original SPN, when discarding the states of the LVs.
However, this procedure is a frequently used approximation for models where MPE is tractable for both model RVs and LVs, but not for model RVs alone.

\begin{figure}
\begin{algorithmic}[1]
\Procedure{MPESelective}{$\SPN, \e$}
\State Initialize zero-vector $\x^*$ of length $|\X|$
\State Evaluate $\e$ in corresponding MPN $\hat\SPN$ (upwards pass)
\State $Q \curvearrowleft $ root node of MPN
\While {$Q$ not empty} 
\State $\hat \Node \curvearrowleft Q$
\If {$\hat \Node$ is a max node}
\State $Q \curvearrowleft \underset{\hat \Child \in \ch(\hat \Node)}{\arg\max} ~ \left \{\w_{\hat\Node,\hat\Child} \, \hat\Child \right \}$
\ElsIf {$\hat \Node$ is a product node}
\State $\forall \hat \Child \in \ch(\hat \Node): Q \curvearrowleft \hat \Child$
\ElsIf {$\hat \Node$ is a maximizing distribution node}
\State $\Node \leftarrow$ corresponding distribution node
\State $\subv{\x^*}{\scope(\Node)} = \underset{\x \in \subv{\e}{\scope(\Node)}}{\arg\max} ~ \Node(\x)$
\EndIf
\EndWhile 
\State \Return $\x^*$
\EndProcedure
\end{algorithmic}
\caption{Pseudo-code for MPE inference in selective SPNs.}
\label{algo:MPESelective} 
\end{figure}

In \cite{Poon2011}, \textproc{MPESelective} was applied to \emph{original} SPNs, not to \emph{augmented} SPNs, but also with the goal to recover an MPE solution over both model RVs and LVs.
The states of the LVs were assigned during max-backtracking, as sum-children and LV states are in one-to-one correspondence.
The states of the LVs whose sums are \emph{not visited} during backtracking, are not assigned --
again, this causes some confusion, since some LVs appear to be undefined in some contexts, cf.~the illustrations in Section~\ref{sec:latentVar}.
However, since this algorithm was used as approximation for MPE over model RVs by discarding the states of the LVs, this situation was not paid any further attention.

Nevertheless, as we show here, applying \textproc{MPESelective} to original (non-selective) SPNs effectively ``simulates'' \textproc{MPESelective} in the corresponding augmented SPN.
Thereby, however, \emph{deterministic} twin-weights are implicitly assumed, i.e.~twin-weights which are 0, except a single 1. 
To see this, let us modify \textproc{MPESelective}, such that it can be applied to an original SPN, 
but returning an MPE solution for the corresponding augmented SPN. 
First note that in the augmented MPN, every twin node simply outputs the maximal twin-weight among all children whose states are contained in evidence $\e$.
For twin node ${\bar \SumNode}$, let this maximal weight be denoted by $\hat \w_{{\bar \SumNode}}$.
The effect of the twin nodes can now be simulated in the original SPN by replacing each weight $\w_{\SumNode,\Child}$ in the original SPN by $\w_{\SumNode,\Child} \times \tilde \w_{\SumNode,\Child}$.
Here $\tilde \w_{\SumNode,\Child}$ is a correction factor and given as $\tilde \w_{\SumNode,\Child} = \prod_{\bar \SumNode} \hat \w_{\bar \SumNode}$, where the product runs over all twins of those sums for which $\SumNode$ is a conditioning sum.
By using these corrected weights, each max node in the corresponding MPN gets the same input as in the MPN of the augmented SPN, i.e.~the twin nodes are simulated.
We can identify the maximizing states of those LVs whose sums are visited during backtracking, as in \cite{Poon2011}.
The states of the sums which are not visited are given by the child which correspond to the maximal twin-weight $\hat \w_{{\bar \SumNode}}$.
Pseudo-code for this somewhat technical modification of \textproc{MPESelective} can be found in \cite{Peharz2015b}.

\begin{figure}
\centering
  \includegraphics[width=0.45\textwidth]{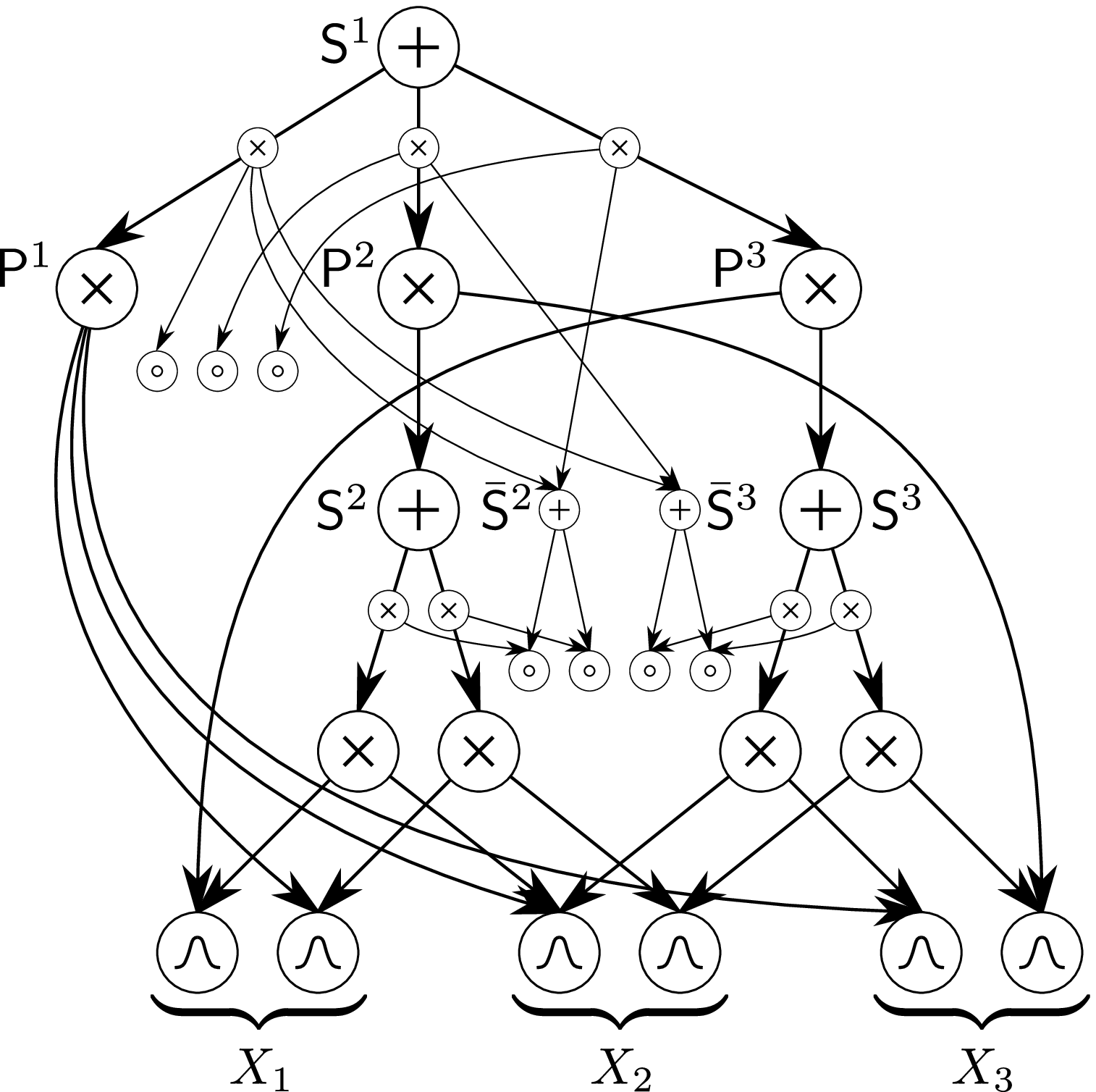}
 \caption{Illustration of the low-depth bias using an SPN over RVs $\{X_1,X_2,X_3\}$. The structure introduced by augmentation is depicted by small nodes and edges. When deterministic twin-weights are used, the state of $Z_{\SumNode^1}$ corresponding to $\ProdNode^1$ is preferred over $\ProdNode^2$ and $\ProdNode^3$, since their probabilities are ``dampened'' by the weights of $\SumNode^2$ and $\SumNode^3$, respectively.}
\label{fig:problemMPE}
\end{figure}

We see that the algorithm used in \cite{Poon2011} is essentially equivalent to \textproc{MPESelective} in 
augmented SPNs when $\tilde \w_{\SumNode, \Child} = 1$ for all sum nodes, which implies that the twin-weights are deterministic.
Therefore, although the LV model in \cite{Poon2011} is not completely specified and it was not shown
that the Viterbi-like algorithm recovers an MPE solution, it nevertheless corresponds to MPE
inference in the augmented SPN for special twin-weights, i.e.~deterministic weights.

However, using deterministic twin-weights is a rather unnatural choice, since this
prefers one arbitrary state over the others in cases where this LV is actually ``rendered irrelevant''. 
In this case, MPE inference also has a bias towards less structured sub-models, which we call \emph{low-depth bias}.
This is illustrated in Fig.~\ref{fig:problemMPE}, which shows an SPN over
three RVs $X_1, X_2, X_3$. 
The augmented SPN has two twin sum nodes $\bar\SumNode^2$ and $\bar\SumNode^3$, corresponding to $\SumNode^2$ and $\SumNode^3$, respectively. 
When their twin-weights are deterministic, the selection of the state of $Z_{\SumNode^1}$
is \emph{biased} towards the state corresponding to $\ProdNode^1$, which is a distribution assuming 
independence among $X_1$, $X_2$ and $X_3$. 
This comes from the fact, that the values of $\ProdNode^2$ and $\ProdNode^3$ are dampened by the weights of 
$\SumNode^2$ and $\SumNode^3$, respectively, which are generally smaller than 1. 
Therefore, when using deterministic weights for twin sum nodes, we introduce a bias towards the 
selection of sub-SPNs that are less deep and less structured. 
Using uniform weights for twin sum nodes is somewhat ``fairer'', 
since in this case 
$\ProdNode^1$ gets dampened by $\bar\SumNode^2$ and $\bar\SumNode^3$,
$\ProdNode^2$ by $\SumNode^2$ and $\bar\SumNode^3$, and 
$\ProdNode^3$ by $\bar\SumNode^2$ and $\SumNode^3$. 
Uniform weights are to some extend the opposite choice to deterministic twin-weights: the
former represent the strongest possible dampening via twin-weights
and therefore actually \emph{penalize} less structured distributions.
Investigating these effects further is subject to future work.

\section{Experiments}

\subsection{Experiments with EM Algorithm}   \label{sec:experimentsEM}
In \cite{Poon2011, PoonSPNCode} SPNs were applied to image data, where a generic
architecture reminiscent to convolutional neural networks was proposed. 
We refer to this architecture as PD architecture. 
Standard EM was not used in experiments for two reasons: 
First, explicitly constructing the proposed structure
and to train it with standard EM is hardly possible with
current hardware, since the number of nodes grows $\mathcal{O}(l^3)$,
where $l$ is the square-length of the modeled image domain in pixels \cite{Peharz2015b}. 
Instead, a sparse hard EM algorithm was used, which virtualizes the PD structure, i.e.~sum and products
are generated on the fly (see \cite{PoonSPNCode} for details). 
Second, using standard EM seemed unsuited to train large and dense SPNs,
either because it is trapped in local optima or due to the
gradient vanishing phenomenon.

In our experiments,\footnote{Code available under \cite{LatentSPNcode}.} we investigated three questions:
\begin{enumerate}
\item
Is our derivation of EM correct, both for complete
and missing data?
\item 
Can the result of hard EM  \cite{Poon2011} be improved by
standard EM?
\item
Given a suited sparse structure, does EM yield a
good solution for parameters?
\end{enumerate}
Question 1) is important since the original derivation contained
an error. Questions 2) and 3) are concerned with the
general applicability of EM for training SPN.

\begin{figure}
\centering
 \subfloat[]{
  \includegraphics[width=0.5 \textwidth]{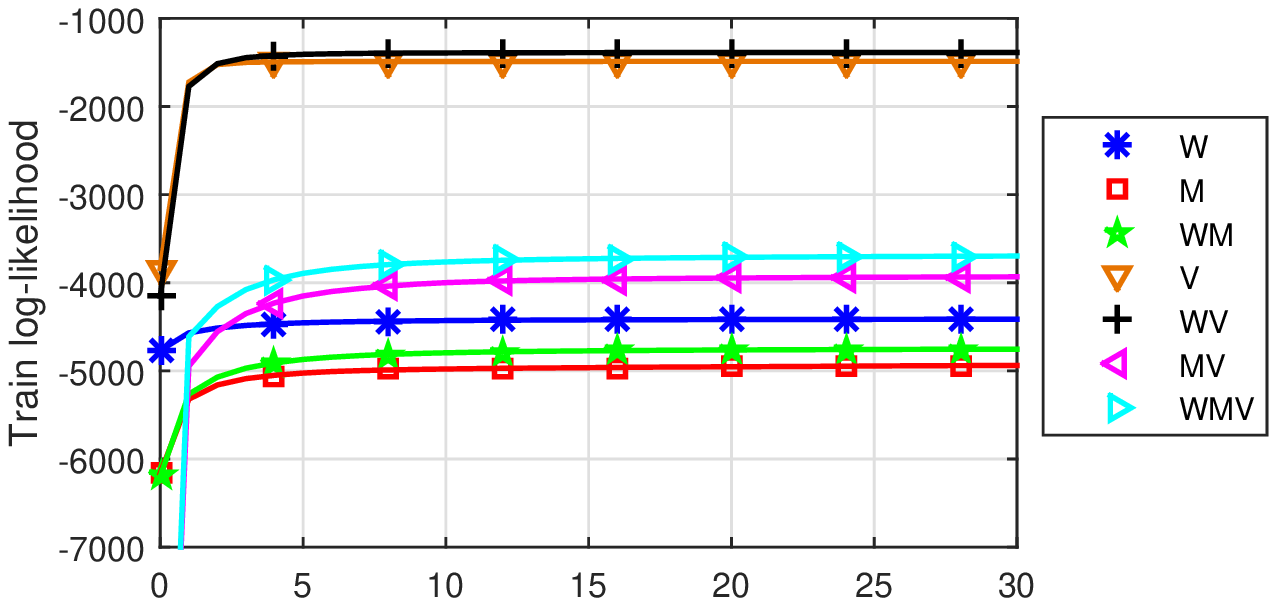}
  \label{fig:EMtrain}  
  }\qquad
 \subfloat[]{
  \includegraphics[width=0.5 \textwidth]{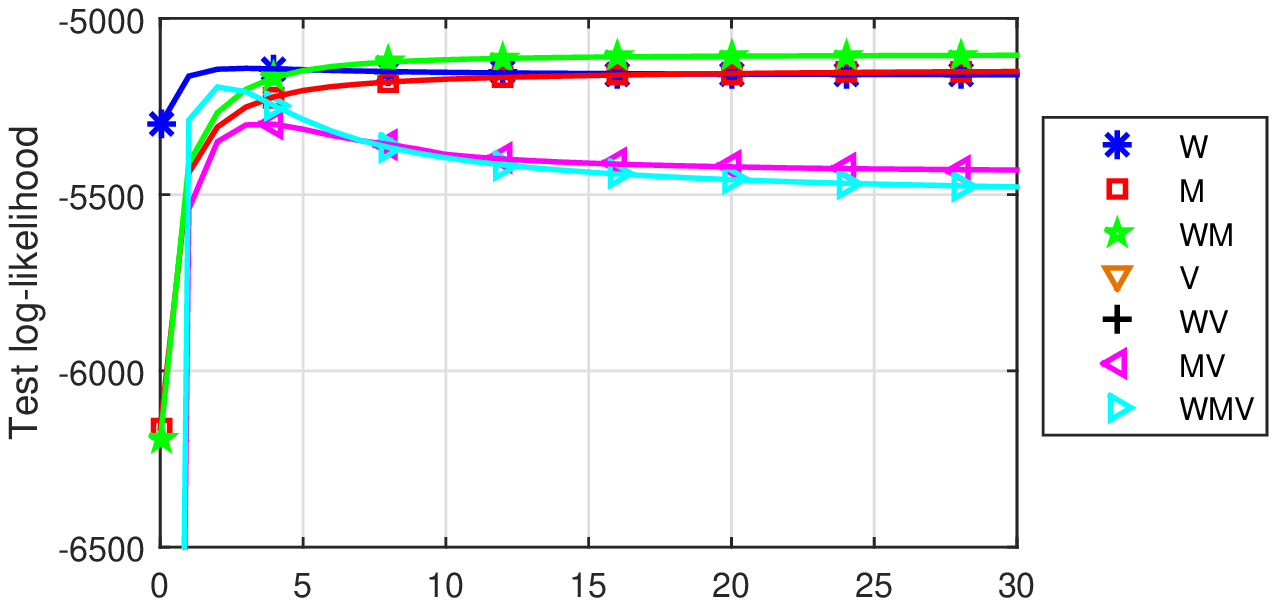}
   \label{fig:EMtest}	
}
\caption{Normalized log-likelihood over EM-iterations, averaged over all 103 datasets and 3 random initializations.
\protect\subref{fig:EMtrain}: Training set.
\protect\subref{fig:EMtest}: Test set; Curves for V and WV are
outside the displayed region, for better readability of the other curves.
They start at approximately $-8000$ nats and decreased to approximately
$-11000$ nats. }
\label{fig:EMresult}
\end{figure}

We used the same datasets and SPN structures as in \cite{Poon2011}, obtainable from \cite{PoonSPNCode}. 
The datasets comprise Caltech-101 (inclusive background class) \cite{FeiFei2007} and the ORL face 
images \cite{Samaria1994}, i.e. in total 103 datasets. 
The input distributions in these SPNs are single-dimensional Gaussians (4 for each pixel), where means were set to the
averages of the 4-quantiles and variances were constantly $1$. 
We ran EM (Fig.~\ref{algo:EM}) for 30 iterations, with various settings:
\begin{itemize}
\item 
Update any combination of the three different types
of parameters, i.e. sum-weights, Gaussian means and
Gaussian variances. 
Each set of parameters types is encoded by a string of 
letters W (weights), M (means) and V (variances). 
(7 combinations)
\item
Use original parameters for initialization, obtained
from \cite{PoonSPNCode}, or use 3 random initialization, where sum-weights
are drawn from a Dirichlet distribution with uniform $\alpha=1$ hyper-parameter 
(i.e. uniform distribution on the standard simplex), Gaussian means
are uniformly drawn from $[-1,1]$ and Gaussian variances from $[0.01,1]$. 
Only parameters which are actually updated are initialized randomly; otherwise
the original parameters \cite{Poon2011} are used and kept fixed.
(4 combinations)
\item
Use complete data or missing training data, randomly discarding $33\%$ or $66\%$ of the observations, independently for each sample. 
(3 combinations)
\end{itemize}
Thus, in total we ran EM $7 \times 4 \times 3 \times 103 = 8652$ times, yielding $259560$ EM-iterations. 
To avoid pathological solutions we used a lower bound of $0.01$ for the Gaussian variances. 
In \emph{no iteration} we observed a decreasing likelihood on the training set,\footnote{Except for tiny occasional decreases (always $< 10^{-8}$) after EM had converged, which can be attributed to numerical artifacts.} i.e. our derived EM algorithm showed monotonicity in our experiments. 
Moreover, as can be seen in Fig.~\ref{fig:EMtrain}, the training log-likelihood actually increased over iterations. 
The curves for the missing data scenarios are similar. 
This gives affirmative evidence for question 1).

Fig.~\ref{fig:EMtest} shows the log-likelihood on the test set. 
Note that optimizing the parameter sets V and WV led to severe overfitting: 
while achieving extremely high likelihoods on the training set, 
they achieved extremely poor likelihoods on the test set. 
Also the parameter sets MV and WMV tend to overfit, although not as strong as V and WV.

Regarding question 2), we closer inspected the test log-likelihood
when the original parameters are used for initialization,
i.e. when the parameters obtained by \cite{PoonSPNCode} are post-trained
using EM. 
Table~\ref{tab:posttrainEM} summarizes the results. 
When parameter sets not including Gaussian variances are optimized
(i.e.~W, M, and WM), the test log-likelihood increased
most of the time, i.e. for $83.5\%$ (M) to up to $92.23\%$ (WM) of the datasets.
Furthermore, having oracle knowledge about the ideal number of iterations 
(i.e. column best), the average log-likelihood increased by $0.58\%$ (M) to up to
$1.39\%$ (WM) relative to the original parameters. 
Most of this improvement happens in the first iteration, yielding $0.52\%$
(M) up to $1.05\%$ (WM) improvement. 
These results indicate that the parameters obtained by \cite{PoonSPNCode} slightly underfit the
given datasets. 
Similar as in Fig.~\ref{fig:EMresult}, we see that parameter sets including the Gaussian variances 
(V, WV, MV, WMV) are prone to overfitting: more than $60\%$ of the datasets
decreased their test log-likelihood during EM. 
However, in the remaining $40\%$ of the datasets, the test log-likelihood could
be improved \emph{substantially} by at least $14\%$ on average.
\begin{table}
\caption{
Changes in test log-likelihoods when original parameters are
post-trained using EM. 
\% inc.: percentage of datasets where log-likelihood increased in the first iteration. 
\% all, \% pos., \% neg.: relative change of log-likelihood, averaged over all datasets, 
datasets with positive change, datasets with negative change, respectively.
}
\label{tab:posttrainEM}
\tabcolsep=0.17cm
\begin{tabular}{|l c| c c c | c c c|}
\hline 
 &       &  \multicolumn{3}{|c|}{after 1st iteration}  & \multicolumn{3}{|c|}{best} \\ 
\hline 
 &  \% inc. &  \% all. & \% pos. & \% neg.  &  \% all & \% pos. & \% neg. \\ 
\hline 
W    &  91.26  &  0.55 & 0.61 & -0.03  &  0.87 & 0.96 & -0.03 \\ 
M    &  83.50  &  0.52 & 0.67 & -0.21  &  0.58 & 0.73 & -0.21 \\ 
WM   &  92.23  &  1.06 & 1.18 & -0.30  &  1.39 & 1.53 & -0.30 \\ 
V    &  39.81  &  -13.47 & 14.44 & -31.93  &  -13.45 & 14.51 & -31.93 \\ 
WV   &  39.81  &  -13.41 & 14.79 & -32.06  &  -13.33 & 14.98 & -32.06 \\ 
MV   &  38.83  &  -17.24 & 14.27 & -37.25  &  -17.21 & 14.35 & -37.25 \\ 
WMV  &  38.83  &  -17.18 & 14.63 & -37.37  &  -17.12 & 14.78 & -37.37 \\ 
\hline 
\end{tabular}
\end{table}

We now turn to question 3). 
As pointed out above, a hard EM variant was used in \cite{Poon2011, PoonSPNCode} which at the same time
finds the effective SPN structure. 
Optimizing W using the 3 random initialization amounts to using the oracle structure
obtained by \cite{Poon2011, PoonSPNCode}, discarding the learned parameters. 
For each dataset we selected the random initialization which yielded the highest 
likelihood on the training set in iteration 30. 
For this run, we compared the log-likelihoods with the log-likelihoods obtained 
by the original parameters. 
The results are summarized in Table~\ref{tab:randinitEM}. 
\begin{table}
\caption{
Log-likelihoods when sum-weights (W) are trained, using random
initialization. 
$\%>$: percentage of data sets, where log-likelihood is larger than for original parameters. 
$\%$ all, $\%$ pos., $\%$ neg.: relative log-likelihood w.r.t.~original parameters, 
for all data sets, data sets where relative log-likelihood is positive/negative, respectively.
}
\label{tab:randinitEM}
\tabcolsep=0.12cm
\begin{tabular}{|l | c c c c | c c c c|}
\hline 
 &      \multicolumn{4}{|c|}{after 1st iteration}  & \multicolumn{4}{|c|}{best} \\ 
\hline 
 &  \%$>$ &  \% all. & \% pos. & \% neg.  & \%$>$ & \% all & \% pos. & \% neg. \\ 
\hline 
train & 70.87 & 0.68 & 1.38 & -1.00 & 100.00 & 3.97 & 3.97 & - \\ 
test & 41.75 & -0.11 & 0.40 & -0.48 & 67.96 & 0.46 & 0.76 & -0.18 \\ 
\hline 
\end{tabular}
\end{table}
We see that on all data
sets the log-likelihood on the training set is larger than
for the original parameters. This is also the case for each
individual random start (not just best one) -- every random
restart always yielded a higher training log-likelihood than
the original parameters. 
Thus, by considering the actual optimization objective 
-- the likelihood on the training set -- EM successfully trains SPNs, 
given a suited oracle structure.
Furthermore, as can be seen in Table~\ref{tab:randinitEM}, 
EM is also not more prone to overfitting than the algorithm in \cite{Poon2011}: 
on $67.96\%$ of the datasets, EM delivered a higher test log-likelihood
than the original parameters, when using oracle knowledge
about the ideal number of iterations (column best).

\subsection{Experiments with MPE Inference}  
To illustrate correctness of \textproc{MPESelective} (Fig.~\ref{algo:MPESelective}) when applied to augmented SPNs, we
generated SPNs using the PD architecture \cite{Poon2011}, arranging
$4$, $9$ and $16$ binary RVs in a $2 \times 2$, $3 \times 3$ and $4 \times 4$ grid,
respectively. 
As inputs we used two indicator variables for each RV representing their two states. 
The sum-weights were drawn from a Dirichlet distribution with uniform $\alpha$-parameters, 
where $\alpha \in \{0.5,1,2\}$. 
For all networks we drew $100$ independent parameters sets. 
We ran \textproc{MPESelective} on the augmented SPN, once equipped with uniform twin-weights and once with deterministic twin-weights.
For uniform twin-weights, we denote the result obtained by \textproc{MPESelective} as \textproc{MPEUni}.
For deterministic twin-weights, we denote the result as \textproc{MPEDet}.
As described in Section~\ref{sec:MPE}, \textproc{MPEDet} corresponds essentially to the result when \textproc{MPESelective} is applied to the original SPN \cite{Poon2011}.
For each assignment, the log-likelihoods were evaluated in the augmented SPN with
deterministic weights, the augmented SPN with uniform weights and in 
the original SPN (discarding the states of the LVs). 
Additionally, we found ground truth MPE assignments in the two augmented SPNs and the 
original SPN using exhaustive enumeration. 
The results relative to the ground truth MPE solutions are shown in 
Tables~\ref{tab:MPESelective}, \ref{tab:MPESelectiveDet}, and \ref{tab:MPEoriginal}. 
As can be seen, \textproc{MPEUni} always finds an MPE solution in the augmented SPN 
with uniform twin-weights and \textproc{MPEDet} always finds an MPE solution
in augmented SPNs with deterministic twin-weights. 
This gives empirical evidence for the correctness of \textproc{MPESelective} for MPE 
inference in augmented SPNs. 

Furthermore, we wanted to investigate the quality of both algorithms
when serving as approximation for MPE inference in the original SPNs. 
For the SPNs considered here, \textproc{MPEDet} delivered on average slightly
better approximations than \textproc{MPEUni}.
However, these results should be interpreted with
caution, due to the rather similar nature of the distributions
considered here. Closer investigating approximate MPE for
(original) SPNs is an interesting direction and will be subject
to future research.
\begin{table}
\center
\caption{
Differences of log-likelihood to the ground-truth MPE solution found by
exhaustive enumeration, averaged over 100 independent draws of
sum-weights. Numbers in parentheses are the number of times where
an MPE solution was found. Results for augmented SPNs using
uniform twin-weights.
}
\label{tab:MPESelective}
\tabcolsep=0.2cm
\begin{tabular}{|l c c c|}
\hline 
 &  & \textproc{MPEDet} & \textproc{MPEUni}  \\ 
\hline 
       & $\alpha=0.5$ & 0.00 (100) & 0.00 (100) \\ 
4 RVs  & $\alpha=1.0$ & 0.00 (100) & 0.00 (100) \\ 
       & $\alpha=2.0$ & 0.00 (100) & 0.00 (100) \\ 
\hline
       & $\alpha=0.5$ & -0.10 (70) & 0.00 (100) \\ 
9 RVs  & $\alpha=1.0$ & -0.10 (68) & 0.00 (100) \\ 
       & $\alpha=2.0$ & -0.11 (62) & 0.00 (100) \\ 
\hline
       & $\alpha=0.5$ & -0.63 (19) & 0.00 (100) \\ 
16 RVs & $\alpha=1.0$ & -0.85 (12) & 0.00 (100) \\ 
       & $\alpha=2.0$ & -0.82 (12) & 0.00 (100) \\ 
\hline
\end{tabular}
\end{table}
\begin{table}
\center
\caption{
Similar as in Table~\ref{tab:MPESelective}. Results for augmented SPNs using deterministic
twin-weights.
}
\label{tab:MPESelectiveDet}
\tabcolsep=0.2cm
\begin{tabular}{|l c c c|}
\hline 
 &  & \textproc{MPEDet} & \textproc{MPEUni}  \\ 
\hline 
        & $\alpha=0.5$ & 0.00 (100) & 0.00 (100) \\ 
4 RVs  & $\alpha=1.0$  & 0.00 (100) & 0.00 (100) \\ 
       & $\alpha=2.0$  & 0.00 (100) & 0.00 (100) \\ 
\hline
       & $\alpha=0.5$  & 0.00 (100) & -0.10 (70) \\ 
9 RVs  & $\alpha=1.0$  & 0.00 (100) & -0.12 (68) \\ 
       & $\alpha=2.0$  & 0.00 (100) & -0.15 (62) \\ 
\hline
       & $\alpha=0.5$  & 0.00 (100) & -0.89 (19) \\ 
16 RVs & $\alpha=1.0$  & 0.00 (100) & -1.11 (12) \\ 
       & $\alpha=2.0$  & 0.00 (100) & -1.01 (12) \\ 
\hline
\end{tabular}
\end{table}
\begin{table}
\center
\caption{
Similar as in Table~\ref{tab:MPESelective}. Results for original SPNs.
}
\label{tab:MPEoriginal}
\tabcolsep=0.2cm
\begin{tabular}{|l c c c|}
\hline 
 &  & \textproc{MPEDet} & \textproc{MPEUni}  \\ 
\hline 
       & $\alpha=0.5$ & -0.06 (72) & -0.06 (72) \\ 
4 RVs  & $\alpha=1.0$ & -0.09 (59) & -0.09 (59) \\ 
       & $\alpha=2.0$ & -0.10 (52) & -0.10 (52) \\ 
\hline
       & $\alpha=0.5$ & -0.31 (32) & -0.38 (27) \\ 
9 RVs  & $\alpha=1.0$ & -0.47 (12) & -0.48 (12) \\ 
       & $\alpha=2.0$ & -0.40 (6)  & -0.37 (7) \\ 
\hline
       & $\alpha=0.5$ & -0.76 (5)  & -1.04 (4) \\ 
16 RVs & $\alpha=1.0$ & -0.76 (3)  & -1.18 (2) \\ 
       & $\alpha=2.0$ & -0.67 (1)  & -0.92 (0) \\ 
\hline
\end{tabular}
\end{table}

\section{Conclusion}
In this paper we revisited the interpretation of SPNs as
hierarchically structured LV model. 
We pointed out that the original approach to explicitly incorporate LVs does not
produce a sound probabilistic model. 
As a remedy we proposed the augmentation of SPNs and proved its soundness as LV model.
Within augmented SPNs, we investigated the independency
structure represented as BN, and showed that the sum-weights
can be interpreted as structured CPTs within this BN. 
Using augmented SPNs, we derived the EM algorithm for sum-weights
and single-dimensional input distributions from
exponential families. 
While MPE-inference is generally NP-hard in SPNs, we showed that a Viterbi-style backtracking algorithm 
recovers an MPE solution in selective SPNs, and in particular in augmented SPNs.
In experiments we give empirical evidence supporting our theoretical results. 
We furthermore showed that standard EM can successfully train generative SPNs, given a suitable
network structure at hand.

\appendices

\section{Proofs}

\subsection{Proof of Proposition \ref{prop:augmentationSound}}
If $\SPN'$ is a complete and decomposable SPN over $\X \cup \Z$, then ${\SPN'}(\X) \equiv \SPN(\X)$ is immediate: 
Computing ${\SPN'}(\x)$ for any $\x \in \val(\X)$ is done by marginalizing $\Z$, i.e.~setting all $\IV{Z_\SumNode}{k} = 1$.
In this case, it is easy to see that none of the structural changes modifies the output of the SPN, i.e.~the outputs of $\SPN$ and $\SPN'$ agree for each $\x$, i.e.~$\SPN'(\X) \equiv \SPN(\X)$.

It remains to show that $\SPN'$ is complete and decomposable, and that the root's scope is $\X \cup \Z$.
Steps \ref{as:introLinksStart}--\ref{as:introLinksEnd} in \textproc{AugmentSPN} introduce the links, representing "private copies" of the sum's children, and clearly leave the SPN complete and decomposable.
In steps \ref{as:introIVStart}--\ref{as:introIVEnd} the LV $Z_\SumNode$ is introduced in the scope of $\SumNode$ and thus in the scope of the root.
Since this is done for all sum nodes, all $\Z$ are introduced in the root's scope.
Steps \ref{as:introIVStart}--\ref{as:introIVEnd} cannot render products non-decomposable, since this would imply that $\SumNode$ is reachable by two distinct children of this product -- a contradiction to the fact that the SPN was decomposable before.
However, as shown in Fig.~\ref{fig:problemAugmentSPNnaiv}, steps \ref{as:introIVStart}--\ref{as:introIVEnd} can render ancestor sums incomplete.
These are treated in steps \ref{as:augmentRemedyStart}--\ref{as:augmentRemedyEnd}.
The twin sum $\bar{\SumNode}$, if introduced, is clearly complete and has scope $\{Z\}$.
Furthermore, incompleteness of any conditioning sum $\SumNode^c$ can only be caused by links not having $Z_\SumNode$ in their scope.
The scope of these links is augmented by $Z_\SumNode$ in step \ref{as:connectTwin}.
These links clearly remain decomposable and moreover, $\SumNode^c$ is rendered complete now.
\hfill $\qed$

\subsection{Proof of Proposition \ref{prop:confspn}}
\textbf{ad \emph{1.)}}
When deleting the IVs and their links, the scopes of any (twin) sum remains the same, since it is complete and is left with one child. 
Thus also the scope of any ancestor remains the same. \\
\textbf{ad \emph{2.)}}
The graph of $\SPN^\y$ is rooted and acyclic, since the root cannot be a link and deleting nodes and edges cannot introduce cycles.
When an IV $\IV{Y}{y}$ is deleted, also the link $\ProdNode_{\SumNode_Y}^{y}$ is deleted, so no internal nodes are left as leaves.
The roots in $\SPN^\y$ and $\SPN'$ are the same, and by point \emph{1.}, $\X \cup \Y \cup \Z$ is the scope of the root.
$\SPN^\y$ is also complete and decomposable:
Whenever an IV and its link are deleted, the corresponding sum node and twin sum node remain trivially complete, since they are left with a single child.
Furthermore, completeness and decomposability of any ancestor of $\SumNode_Y$ or $\bar \SumNode_Y$ is left intact, since neither $\SumNode_Y$ nor $\bar \SumNode_Y$ changes its scope. \\
\textbf{ad \emph{3.)}}
According to point \emph{1.}, the scope of $\Node$ is the same in $\SPN'$ and $\SPN^\y$.
Since $\scope(\Node) \cap \Y = \emptyset$, the disconnected IVs and deleted links are no descendants of $\Node$, i.e.~no descendants of $\Node$ are disconnected during configuration.
Since $\Node$ is present in $\SPN^\y$, it must still be reachable from the root.
Therefore also all descendants of $\Node$ are reachable, i.e. $\SPN^\y_\Node = \SPN'_\Node$. \\
\textbf{ad \emph{4.)}}
When the input is fixed to $\x, \z, \y$, all IVs and links which are deleted from the configured SPN $\SPN^\y$ evaluate to zero in the augmented SPN $\SPN'$.
The outputs of all sums and twin sums are therefore the same in $\SPN'$ and $\SPN^\y$. 
Therefore, also the output of all other nodes remains the same.
This includes the root and therefore $\SPN^\y(\x,\z,\y) = \SPN'(\x,\z, \y)$, for any $\x, \z$.

When $\y' \not= \y$, then there must be a $Y \in \Y$ such that the IV $\IV{Y}{\subv{\y'}{Y}}$ has been deleted, i.e.~$\IV{Y}{\subv{\y'}{Y}} \notin \desc(\Node)$, where $\Node$ is the root of $\SPN^\y$.
Using Lemma~1 in \cite{Peharz2015}, it follows that $\SPN^\y(\x,\z,\y') = 0$.
 \hfill $\qed$

\subsection{Proof of Lemma \ref{lem:singlepath}}
$\SPN^{\z}$ must contain either $\SumNode$ or $\bar{\SumNode}$, since $Z_\SumNode$ is in the scope of the root by Proposition~\ref{prop:confspn}.
To show that \emph{not both} are in $\SPN^{\z}$, let $\bm{\Pi}_k$ denote the set of paths of length $k$ from the root to any node $\Node$ with $Z_\SumNode \in \scope(\Node)$.
For $k > 1$, all paths in $\bm{\Pi}_k$ can be constructed by extending each path in $\bm{\Pi}_{k-1}$ with each child of this path's last node, if it has $Z_\SumNode$ in its scope.
Let $K$ be the smallest number such that there is a path in $\bm{\Pi}_k$ containing $\SumNode$ or $\bar{\SumNode}$.

We show by induction, that $|\bm{\Pi}_k| = 1$, $k=1,\dots,K$.
Note that $\bm{\Pi}_1$ contains a single path $(\Node)$, where $\Node$ is the root, therefore the induction basis holds.

For the induction step, we show that given $|\bm{\Pi}_{k-1}| = 1$, then also $|\bm{\Pi}_{k}| = 1$.
Let $(\Node_1, \dots, \Node_{k-1})$ be the single path in $\bm{\Pi}_{k-1}$.
If $\Node_{k-1}$ is a product node, then it has a single child $\Child$ with $Z_\SumNode \in \scope(\Child)$, due to decomposability.
If $\Node_{k-1}$ is a sum node, then it must be in $\Sanc(\SumNode) \setminus \{\SumNode\}$, and therefore has a single child in the configured SPN. 
Therefore, there is a single way to extend the path and therefore $|\bm{\Pi}_k|=1, k=1,\dots,K$.
This single path does either lead to $\SumNode$ or $\bar{\SumNode}$.
Since $\SumNode \notin \desc(\bar{\SumNode})$ and $\bar{\SumNode} \notin \desc({\SumNode})$, $\SPN^{\z}$ contains a single path to one of them, but not to both.
 \hfill $\qed$

\subsection{Proof of Theorem \ref{theo:conditionalIndependence}}
By Lemma~\ref{lem:singlepath}, for each $\z \in \val(\Z_p)$ the configured SPN 
$\SPN^{\z}$ contains either $\SumNode$ or $\bar{\SumNode}$, but not both.
Let $\ZS$ be the subset of $\val(\Z_p)$ such that $\SumNode$ is in $\SPN^{\z}$ 
and $\bar \ZS$ be the subset of $\val(\Z_p)$ such that $\bar{\SumNode}$ is in 
$\SPN^{\z}$.

Fix $Z_\SumNode = k$ and $\z \in \bm{\mathcal{Z}}$.
We want to compute $\SPN'(Z_\SumNode = k, \Y_n, \z)$, i.e.~we marginalize $\Y_c$.
According to Proposition~\ref{prop:confspn} (\emph{4.}), this equals 
$\SPN^{\z}(Z_\SumNode = k, \Y_n, \z)$.
According to Proposition~\ref{prop:confspn} (\emph{3.}), the sub-SPN rooted at 
former child $\Child^k_\SumNode$ is the same in $\SPN'$ and $\SPN^{\z}$.
Since $\SPN'$ is locally normalized, this sub-SPN is also locally normalized in 
$\SPN^{\z}$.
Since the scope of the former child $\Child^k_\SumNode$ is a sub-set of $\Y_c$, 
which is marginalized, and $\IV{Z_\SumNode}{k} = 1$, the link  $\ProdNode^{k}_{\SumNode}$ 
outputs $1$.
Since $\IV{Z_\SumNode}{k'} = 0$ for $k' \not= k$, the sum $\SumNode$ outputs $\w_k$.

Now consider the set of nodes in $\SPN^\z$ which have $Z_\SumNode$ in their scope, 
not including $\IV{Z_\SumNode}{k}$ and $\ProdNode^k_\SumNode$.
Clearly, since $\bar{\SumNode}$ is not in $\SPN^\z$, this set must be 
$\anc(\SumNode)$.
Let $\Node_1,\dots,\Node_L$ be a topologically ordered list of 
$\anc(\SumNode)$, where $\SumNode$ is $\Node_1$ and $\Node_L$ is the root.
Let $\Y_{n,l} := \scope(\Node_l) \cap \Y_n$ and 
$\Z_l := \scope(\Node_l) \cap \Z_p$.
We show by induction that for $l=1,\dots,L$, we have
\begin{equation}
\label{eq:factor_induct}
 \Node_l(Z_\SumNode = k, \Y_{n,l}, \subv{\z}{\Z_l}) = \w_k \, \Node_l(\Y_{n,l}, \subv{\z}{\Z_l}).
\end{equation}
Since $\Y_{n,1} = \emptyset$ and ${\Z_1} = \emptyset$, and 
$\Node_1(Z_\SumNode = k) = \w_k$, the induction basis holds.
Assume that \eqref{eq:factor_induct} holds for all $\Node_1,\dots,\Node_{l-1}$.
If $\Node_l$ is a sum, we have due to completeness
\begin{align}
 \Node_l(Z_\SumNode = k, {\Y_{n,l}}, \subv{\z}{\Z_l}) 
&= \sum_{\Child \in \ch(\Node_l)} \w_{\Node_l, \Child} \, \w_k \, \Child({\Y_{n,l}}, \subv{\z}{\Z_l}) \\
 &= \w_k \, \Node_l({\Y_{n,l}}, \subv{\z}{\Z_l}),
\end{align}
i.e.~the induction step holds for sums.
When $\Node_l$ is a product, due to decomposability, it must have a single
child with $Z_\SumNode$ in its scope.
Hence, this child must be a node $\Node_m \in \anc(\SumNode)$
We have
\begin{align}
& \Node_l(Z_\SumNode = k, \Y_{n,l}, \subv{\z}{\Z_l}) \\
&= \w_k \, \Node_m(\Y_{n,m}, \subv{\z}{\Z_m})  \prod_{\Child \in \ch(\Node_l) \setminus \Node_m} \Child(\Y_{n,l} \cap \scope(\Child)) \\
&= \w_k \, \Node_l({\Y_{n,l}}, \subv{\z}{\Z_l}),
\end{align}
i.e.~the induction step holds for products.
Therefore, by induction, \eqref{eq:factor_induct} also holds for the root, 
and \eqref{eq:CIaugSum} follows.

Now we show \eqref{eq:CIaugTwinSum}.
If the twin sum $\bar \SumNode$ does not exist, $\bar \ZS$ is empty and \eqref{eq:CIaugTwinSum} holds trivially.
Otherwise, fix the input to $Z_\SumNode = k$ and $\z \in \bar \ZS$.
Clearly, $\bar{\SumNode}$ outputs $\bar{\w}_k$ and \eqref{eq:CIaugTwinSum} can be shown in similar way as 
\eqref{eq:CIaugSum}.
\hfill $\qed$

\subsection{Proof of Theorem \ref{theo:MPEaug}}
We prove the theorem using an inductive argument.
The theorem clearly holds for any $\hat{\DistNode}$ by definition.
Consider a product $\hat{\ProdNode}$ and assume the theorem holds for all $\ch(\hat{\ProdNode})$.
Then the theorem also holds for $\hat{\ProdNode}$, since 
\begin{align}
 \hat{\ProdNode}(\XS)  
= \prod_{{\Child} \in \ch({\hat \ProdNode})} \underset{\x \in \XS}{\max}~\Child(\x)  
= \underset{\x \in \XS} \max  \prod_ {{\Child} \in \ch({\hat \ProdNode})} \Child(\x)  
= \underset{\x \in \XS} \max ~ {\ProdNode}(\x), 
\end{align}
where the max and the product can be switched due to decomposability.

Now consider a max node $\hat{\SumNode}$ and its corresponding sum node $\SumNode$.
Let the \emph{support} of an SPN-node $\Node$ be the set $\sup_\Node := \{\x \cbar \Node(\x) > 0\}$.
Since $\SumNode$ is selective, its support is partitioned by the supports of its children, i.e.
$\sup_\SumNode = \bigcup_{\Child \in \ch(\SumNode)} \sup_\Child$, $\sup_{\Child'} \bigcap \sup_{\Child''} = \emptyset$, for $\Child' \not= \Child''$.
Assuming that the theorem holds for all $\ch(\hat{\SumNode})$, we have
\begin{align}
 \hat\SumNode(\XS)  
&= \underset{\Child \in \ch(\SumNode)}{\max} \w_{\SumNode, \Child} ~ \underset{\x \in \XS}{\max}~\Child(\x)     \\
&= \underset{\Child \in \ch(\SumNode)}{\max} \w_{\SumNode, \Child} ~ \underset{\x \in \sup_\Child \cap \XS}{\max}~\Child(\x)    \label{eq:maxoversupp} \\
&= \underset{\Child \in \ch(\SumNode)}{\max} ~ \underset{\x \in \sup_\Child \cap \XS}{\max}~\w_{\SumNode, \Child} \, \Child(\x)  \label{eq:selSumisChild} \\ 
&= \underset{\x \in \sup_\SumNode \cap \XS}{\max} ~ \SumNode(\x) = \underset{\x \in \XS}{\max} ~ \SumNode(\x).         
\end{align}
In \eqref{eq:maxoversupp} we have a slight abuse of notation, as we actually should use suprema over the sets $\sup_\Child \cap \XS$ and define the supremum over the empty set as $0$.
In \eqref{eq:selSumisChild} we used the fact that the support of the sum node is partitioned by the supports of its children and that for selective sums we have
$\SumNode = \w_{\SumNode, \Child} \, \Child$ whenever we have single child with $\Child > 0$.

We see that the induction step also holds for $\hat \SumNode$.
Therefore, the theorem holds for all nodes. 
\hfill $\qed$

\ifCLASSOPTIONcompsoc
  \section*{Acknowledgments}
\else
  \section*{Acknowledgment}
\fi
We would like to thank the anonymous reviewers for their constructive comments.
This work was supported by the \textbf{Austrian Science Fund (FWF): P25244-N15} and \textbf{Austrian Science Fund (FWF): P27803-N15}.
This research was partly funded by \textbf{ONR grant N00014-16-1-2697} and \textbf{AFRL contract FA8750-13-2-0019}.

\ifCLASSOPTIONcaptionsoff
  \newpage
\fi

\bibliographystyle{IEEEtran}
\bibliography{bibliography}

\begin{IEEEbiography}[{\includegraphics[width=1in,height=1.25in,clip,keepaspectratio]{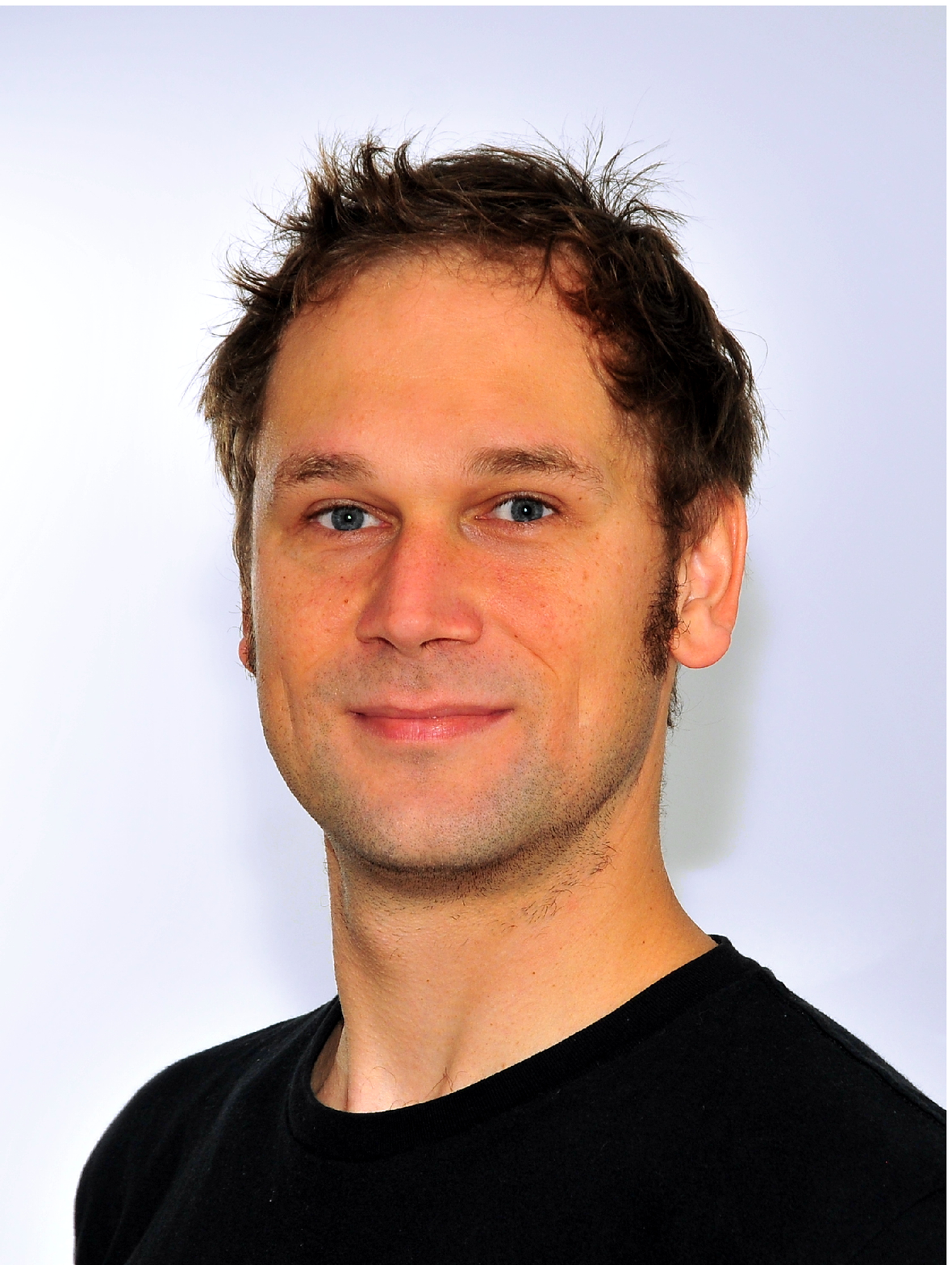}}]{Robert Peharz}
received his MSc degree in Computer Engineering and his Ph.D degree in Electrical Engineering from Graz University of Technology.
His main research interest lies in machine learning, in particular probabilistic modeling, with applications to signal processing,
speech and audio processing, and computer vision.
Currently, he is with the research unit iDN (interdisciplinary developmental neuroscience) at the Medical University of 
Graz, applying machine learning techniques to detect early markers of neurological conditions in infants. 
He is funded by the BioTechMed-Graz cooperation, an interdisciplinary network of the 3 major universities in Graz with a focus 
on basic bio-medical research, technological development and medical applications.
\end{IEEEbiography}

\begin{IEEEbiography}[{\includegraphics[width=1in,height=1.25in,clip,keepaspectratio]{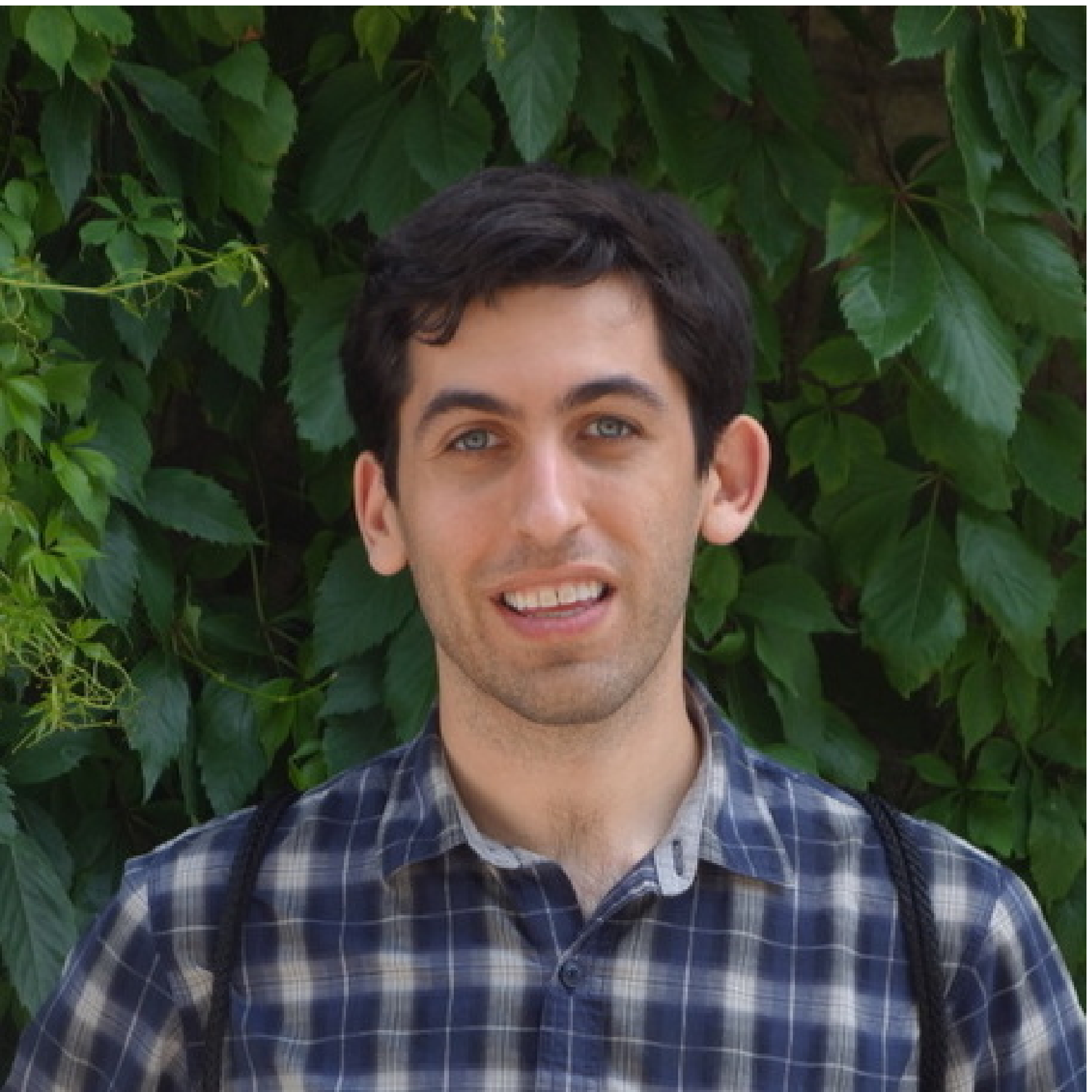}}]{Robert Gens}
 received the S.B. degree in electrical engineering and computer science from the Massachusetts Institute of Technology, Cambridge, MA, USA, in 2009 and the M.Sc. degree in computer science and engineering from the University of Washington, Seattle, WA, USA, in 2012.
During the Summer of 2014, he was a Research Intern at Microsoft Research, Redmond, WA, USA.  
He is currently a Ph.D. student of computer science and engineering at the University of Washington, Seattle, WA, USA.  
He is supported by the 2014 Google Ph.D. Fellowship in Deep Learning.
Mr. Gens was the recipient of an Outstanding Student Paper Award at the Neural Information Processing Systems conference in 2012.
\end{IEEEbiography}

\begin{IEEEbiography}[{\includegraphics[width=1in,height=1.25in,clip,keepaspectratio]{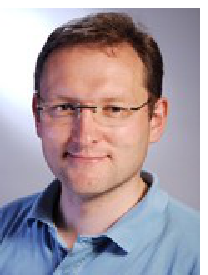}}]{Franz Pernkopf}
received his MSc (Dipl. Ing.) degree in Electrical Engineering at Graz
University of Technology, Austria, in summer
1999. He earned a Ph.D degree from the University of Leoben, Austria, in
2002. In 2002 he was awarded the Erwin Schr\"{o}dinger Fellowship. He
was a Research Associate in the Department of Electrical Engineering at
the University of Washington, Seattle, from
2004 to 2006. Currently, he is Associate Professor at the Laboratory of
Signal Processing and Speech Communication, Graz University
of Technology, Austria. His research interests include machine learning,
discriminative learning, graphical models, feature selection, finite
mixture models, and image- and speech processing applications.
\end{IEEEbiography}

\begin{IEEEbiography}[{\includegraphics[width=1in,height=1.25in,clip,keepaspectratio]{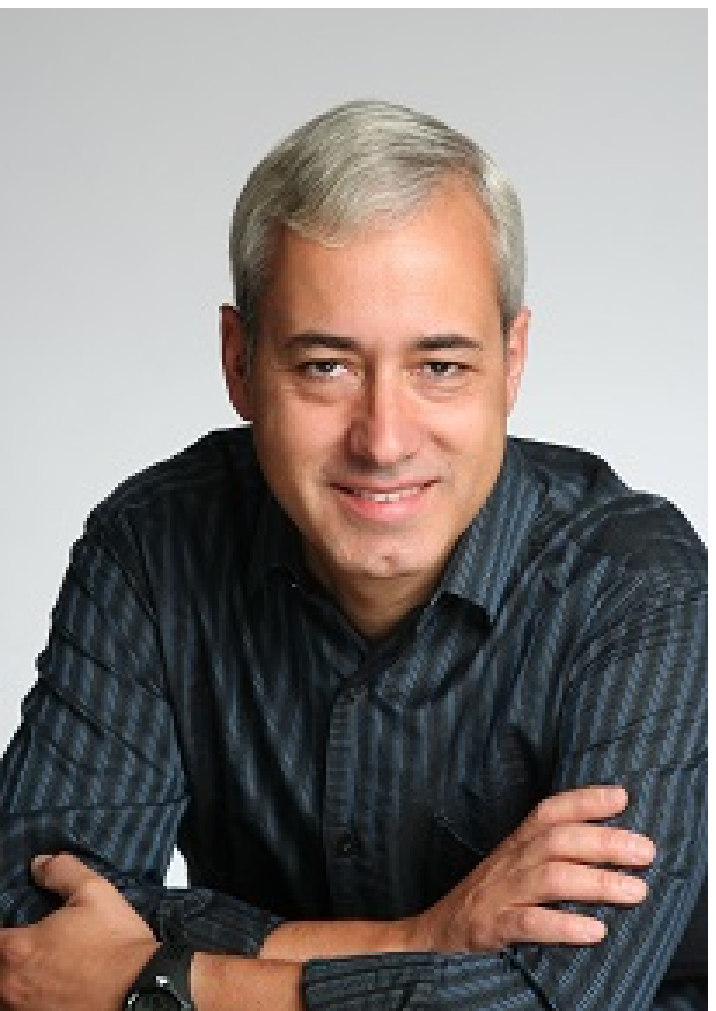}}]{Pedro Domingos}
is a professor of computer science at the University of Washington and the author of “The Master Algorithm”. He is a winner of the SIGKDD Innovation Award, the highest honor in data science. He is a Fellow of the Association for the Advancement of Artificial Intelligence, and has received a Fulbright Scholarship, a Sloan Fellowship, the National Science Foundation’s CAREER Award, and numerous best paper awards. He received his Ph.D. from the University of California at Irvine and is the author or co-author of over 200 technical publications. He has held visiting positions at Stanford, Carnegie Mellon, and MIT. He co-founded the International Machine Learning Society in 2001. His research spans a wide variety of topics in machine learning, artificial intelligence, and data science, including scaling learning algorithms to big data, maximizing word of mouth in social networks, unifying logic and probability, and deep learning.
\end{IEEEbiography}

\end{document}